%% file: paper.tex
\documentclass{article}
\expandafter\def\csname ver@mathpazo.sty\endcsname{}
\usepackage[preprint]{colm2026_conference}
\usepackage{times}

\usepackage{graphicx}
\usepackage{amsmath}
\usepackage{amssymb}
\usepackage{xcolor}
\usepackage{array}
\usepackage{enumitem}
\usepackage{wrapfig}
\usepackage{lineno}
\newcolumntype{L}[1]{>{\raggedright\arraybackslash}p{#1}}

\usepackage{microtype}
\usepackage{hyperref}
\usepackage{url}
\usepackage{booktabs}

\definecolor{darkblue}{rgb}{0, 0, 0.5}
\hypersetup{colorlinks=true, citecolor=darkblue, linkcolor=darkblue, urlcolor=darkblue}

\definecolor{matchgreen}{HTML}{2E7D32}
\definecolor{phantomred}{HTML}{C62828}
\definecolor{verdictblue}{HTML}{1565C0}
\definecolor{decisiveorange}{HTML}{E65100}
\definecolor{rebuttalpurple}{HTML}{6A1B9A}
\definecolor{panelbg}{HTML}{FAFAFA}
\definecolor{panelborder}{HTML}{BDBDBD}

\newcommand{\badge}[2]{%
  \begingroup
  \setlength{\fboxsep}{1.5pt}%
  \fcolorbox{#1!46}{#1!14}{{\color{#1!90!black}\bfseries\fontsize{6.1}{6.8}\selectfont #2}}%
  \endgroup
}

\definecolor{ladderblue}{HTML}{1565C0}
\newcommand{\lev}[1]{%
  \begingroup
  \setlength{\fboxsep}{1.7pt}%
  \raisebox{0.15ex}{\fcolorbox{ladderblue!35}{ladderblue!14}{{\color{ladderblue!92!black}\bfseries\scriptsize L#1}}}%
  \endgroup
}

\title{What Makes a Good AI Review?\\Concern-Level Diagnostics for AI Peer Review}

\author{Ming Jin \\
Bradley Department of Electrical \& Computer Engineering \\
Virginia Tech}

\begin{document}
\ifcolmsubmission
\linenumbers
\fi

\maketitle

\begin{abstract}
Evaluating AI-generated reviews by verdict agreement is widely recognized as
insufficient, yet current alternatives rarely audit which concerns a system
identifies, how it prioritizes them, or whether those priorities align with the
review rationale that shaped the final assessment.
We propose \emph{concern alignment}, a diagnostic framework that evaluates AI reviews at the concern level rather than only at the verdict level.
The framework's core data structure is the \emph{match graph}, a bipartite alignment between official and AI-generated concerns annotated with match type, severity, and post-rebuttal treatment.
From this artifact we derive an \emph{evaluation ladder} that moves from binary accuracy to concern detection, verdict-stratified behavior, decision-aware calibration, and rebuttal-aware decomposition.
In a pilot study of four public AI review systems evaluated in six configurations, concern-level analysis suggests that detection alone does not determine review quality; calibration is often the binding constraint.
Systems detect non-trivial fractions of official concerns yet most mark 25--55\% of concerns on accepted papers as decisive, where, under our operationalization, no official concern on accepted papers was treated as a decisive blocker.
Identical overall verdict accuracy can conceal reject-heavy behavior versus low-recall profiles, and low full-review false decisive rates can partly reflect concern dilution rather than calibrated prioritization.
Most systems do not emit a native accept/reject, and inferring it from review tone is method-sensitive, reinforcing the need for concern-level diagnostics that remain stable across inference choices.
The contribution is a reusable evaluation framework for auditing which
concerns AI reviewers identify, how they weight them, and whether those
priorities align with the review rationale that informed the paper's final assessment.
\end{abstract}

\section{Introduction}
\label{sec:intro}

A good AI review identifies the right concerns, assigns them the right weight, and aligns with the rationale that actually decides the paper.
Most AI review systems are evaluated primarily by verdict agreement or coarse similarity to human reviews \citep{liang2024can,lu2024aiscientist,darcy2024marg,oar2024,gao2025reviewagents}.
Recent work has started to move beyond verdict matching by comparing attention over review facets \citep{shin2025blindspots}, scoring review quality dimensions \citep{garg2025revieweval}, checking premise-level factuality \citep{ryu2025reviewscore}, and benchmarking limitation identification \citep{xu2025limitgen}.
But researchers do not act on verdicts or facet histograms alone; they act on
concrete concerns and on how much those concerns should matter.

Verdict-only evaluation misses the structure that makes reviews useful.
A system can reach moderate verdict accuracy by rejecting almost everything
(Appendix~\ref{app:verdict_split}), yet that behavior is invisible in overall
accuracy. It can recover only a small fraction of the concerns that actually
drove the decision (Table~\ref{tab:core_metrics}), or it can recover the
right concern family but assign it materially different severity
(Figure~\ref{fig:matchgraph}; Figure~\ref{fig:fdr_bar}). These are
substantively different failure modes, but overall accuracy compresses them
into one number. It cannot tell whether a system reached the right verdict
for the wrong reasons, surfaced the right concerns but miscalibrated their
decision weight, or flagged concerns that the official process ultimately
treated as non-blocking. Because reviews are consumed as prioritized concern
lists, evaluation that discards concern-level structure discards the unit
researchers actually use.

We introduce \textbf{concern alignment}, a diagnostic framework organized as an \emph{evaluation ladder}.
Level~0 (\lev{0}) asks whether the verdict is right; Level~1 (\lev{1}) whether the system surfaces the same concerns the official reviewers raised; Level~2 (\lev{2}) whether its behavior changes across accepted and rejected papers; Level~3 (\lev{3}) whether it assigns decision weight appropriately; and Level~4 (\lev{4}) whether it attends to the concerns that mattered most for the decision.
The ladder's central artifact is the \emph{match graph}: a bipartite alignment between official and AI-generated concerns annotated with match type, severity, and the area chair's post-rebuttal treatment.
All primary metrics derive from this single, auditable artifact.

The pilot's main empirical lesson is that detection alone does not determine review quality; calibration is often the binding constraint.
A system can notice a real weakness yet still be wrong about the paper because it overstates how much that weakness should count.
It can also produce a long, weakly prioritized concern list that eventually overlaps official concerns without telling an author which issues most deserve revision effort first.
These are failures of selective attention, not just of detection.

We demonstrate the framework on four public systems (System~L \citep{liang2024can}, System~A \citep{lu2024aiscientist}, System~O \citep{oar2024}, and System~M \citep{darcy2024marg}), across papers from three major ML venues.
Concern-level analysis reveals that identical verdict accuracy can conceal reject-heavy and low-recall profiles; that most single-agent systems mark 25--55\% of concerns on accepted papers as decisive; that changing the model while holding prompts fixed shifts reviewer behavior in measurable, non-uniform ways; and that some systems attend as readily to resolved concerns as to decisive blockers~(\S\ref{sec:findings}).

\section{Concern Alignment Framework}
\label{sec:framework}

\subsection{Design Principles}
\label{sec:problem}

We organize the framework around three criteria for a useful review. These
are modeling assumptions; readers may reasonably weight them differently. We
state them to make the evaluation's design ground explicit.

\textbf{Prioritization.} Useful reviews distinguish blocking concerns from improvement suggestions. When decisive blockers are present, they should be identifiable; when they are absent, real but non-blocking issues should not be inflated into blockers.
The distinction between ``this must be fixed for the paper to be acceptable'' and ``this would improve the paper'' is often the most consequential judgment in the review.
Appendix~\ref{app:severity_examples} provides worked examples of severity determination for each system, and Appendix~\ref{app:severity_policy} lists representative decisive and non-decisive concerns to support independent calibration of this judgment.

\textbf{Distinguishing power.} Useful reviews should change their concern profile with paper quality. If a system raises the same kinds of concerns, at the same severity, on accepted and rejected papers, its feedback provides weak guidance about which issues most deserve revision effort.

\textbf{Aligned coverage.} A useful AI review should recover the concerns that informed the official review rationale, may surface additional valid concerns the official reviewers did not mention, and should avoid generic, tangential, or unsupported complaints.

These criteria decompose into measurable sub-properties, detection and
calibration, tested by the evaluation ladder (\S\ref{sec:ladder}).
Verdict and concern-level analysis are complementary rather than competing signals: the verdict provides the directional label, while concerns provide the diagnostic decomposition.
Conditioning concern analysis on the verdict (\lev{2} and above) reveals, for example, that a system with moderate overall accuracy rejects nearly every accepted paper, a behavior invisible to both binary accuracy and verdict-blind recall.

\subsection{Match Graphs}
\label{sec:matchgraph}

The framework uses area chair (AC) decisions as an operational anchor for
calibrating concern severity and decision weight. Human decisions are
themselves noisy~\citep{cortes2021inconsistency,beygelzimer2023neurips}, and
some accepted papers may contain genuine blockers that were deprioritized.
We adopt this anchor because post-deliberation AC judgments, informed by
reviews, rebuttals, and discussion, provide the best available proxy for
how each concern was ultimately weighted. The assumption is conservative:
it gives a concrete reference point for measuring calibration while
acknowledging its limitations (\S\ref{sec:discussion}).

\begin{figure*}[t]
\centering
\includegraphics[width=0.98\textwidth]{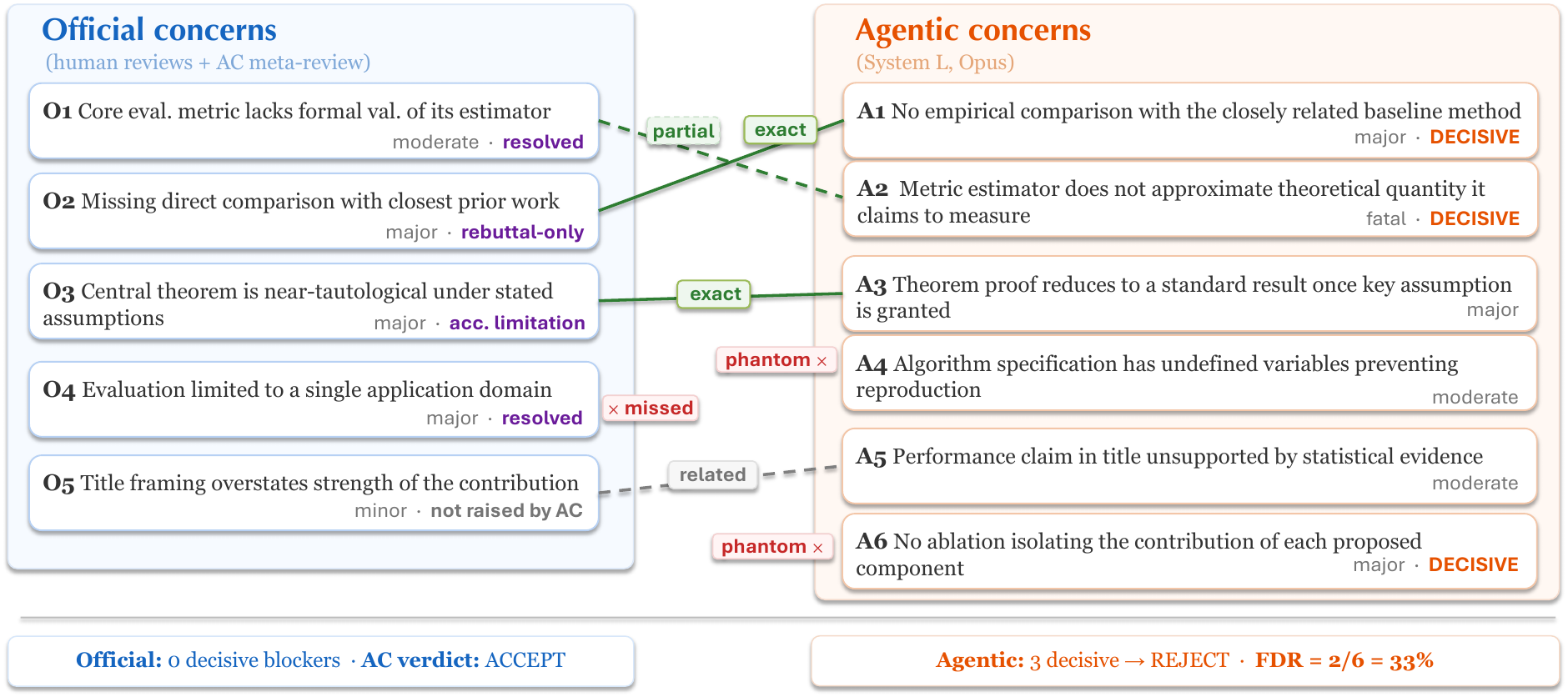}
\caption{Match graph for an accepted paper. Official concerns (left) carry severity labels and AC treatment badges; agentic concerns (right) carry severity and decisive flags. Edges show match type: \texttt{exact} and \texttt{partial} count for strict metrics; \texttt{related} is excluded. Unmatched official concerns are \emph{misses}; unmatched agentic concerns are \emph{phantoms}. Two of six agentic concerns are false decisive flags (33\%); the third decisive flag (A1) correctly identifies a weakness whose fix is not visible in the reviewed PDF.}
\label{fig:matchgraph}
\end{figure*}

A \emph{match graph} is a bipartite alignment between official concerns $\mathcal{O} = \{o_1, \ldots, o_m\}$ (from human reviews, rebuttal, and meta-review) and agentic concerns $\mathcal{A} = \{a_1, \ldots, a_n\}$ (from the AI review) for a single paper.
Each official concern carries severity $s_i \in \{\text{fatal}, \text{major}, \text{moderate}, \text{minor}\}$ and an \emph{AC treatment} label recording the area chair's post-rebuttal disposition (decisive blocker, unresolved, accepted limitation, resolved, dismissed, reframed feature, or not mentioned when the AC does not address the concern). Resolved concerns additionally carry an \texttt{addressed\_in\_pdf} flag; those marked resolved whose fix had not yet appeared in the reviewed PDF version are called ``rebuttal-only'' in figure callouts and remain valid detection targets for the PDF-facing AI reviewer.
Each agentic concern carries severity and a \texttt{decisive} flag.
An edge $e_{ij}$ connects $o_i$ to $a_j$ when both address the same issue: \texttt{exact} if fixing one fixes the other, \texttt{partial} if they share an issue family but differ in scope, \texttt{related} if topically nearby but distinct (excluded from strict metrics).
Each edge is annotated with severity alignment under a hybrid tolerance rule: fatal requires an exact match, while one-level gaps among non-fatal concerns count as matches and larger gaps are coded as under/over.
Unmatched official concerns are \emph{misses}; unmatched agentic concerns are \emph{phantoms} (Figure~\ref{fig:matchgraph}).

Match graphs are constructed in five steps: (1)~official concern extraction from OpenReview with severity, AC treatment, and decisive flags; (2)~agentic concern extraction, deduplicated across review sections; (3)~bipartite matching via scope test; (4)~semantic verification by an independent auditor with 32 calibration exemplars (Appendix~\ref{app:exemplars}); (5)~metric derivation.
Three design choices are central: using AC treatment as an operational anchor
enables rebuttal-aware metrics (under the assumption that post-deliberation
AC judgments, while noisy, provide the best available proxy for how each
concern was ultimately weighted); only exact and partial edges count for
strict metrics; and severity uses the same hybrid tolerance rule throughout
the paper (Appendix~\ref{app:protocol}).

\subsection{The Evaluation Ladder}
\label{sec:ladder}

The ladder has five levels, grouped into three broader stages: verdict agreement (\lev{0}), concern-set alignment (\lev{1}--\lev{2}), and decision-weight calibration (\lev{3}--\lev{4}).
Each level answers a diagnostic question the level below cannot.
Figure~\ref{fig:ladder} illustrates the progression on a single accepted-paper review; formal definitions appear in Appendix~\ref{app:metrics}.

\begin{figure*}[t]
\centering
\includegraphics[trim=21pt 22pt 24pt 9pt,clip,width=0.98\textwidth]{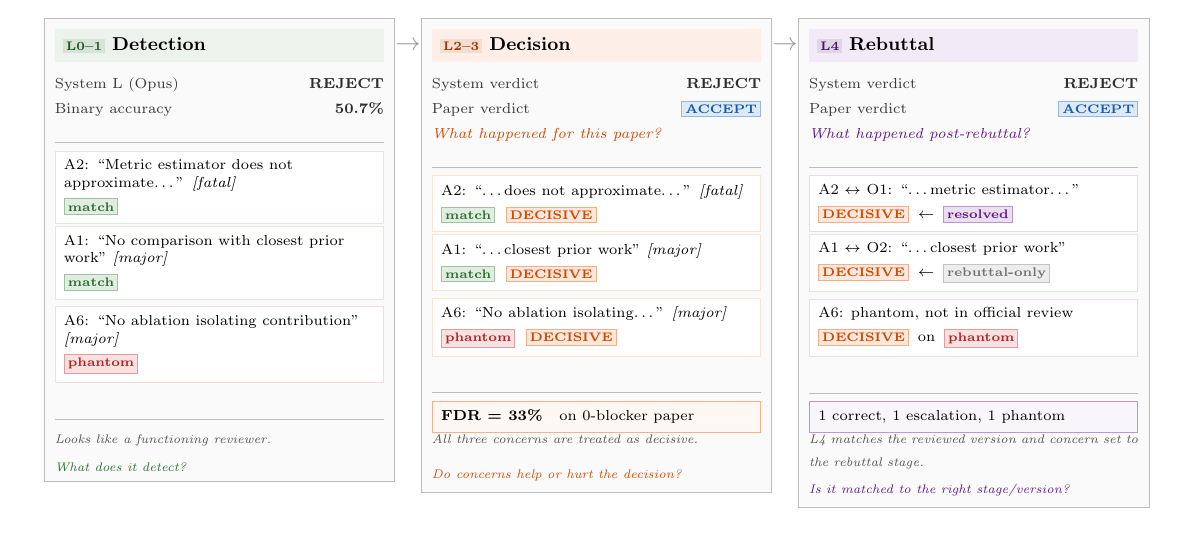}
\vspace{-5pt}
\caption{Progressive reveal: the same AI review (System~L, Opus, accepted paper) diagnosed at increasing depth.
\textbf{Left}: looks like a functioning reviewer.
\textbf{Center}: All three shown concerns carry decisive flags on a paper with zero blockers (FDR~=~33\%).
\textbf{Right}: the evaluation must match the reviewed paper version and concern set to the rebuttal stage. A2 escalates a \emph{resolved} concern (calibration failure); A1 correctly detects a weakness whose fix never appeared in the reviewed PDF (\emph{rebuttal-only}); A6 is a fabricated phantom.}
\label{fig:ladder}
\end{figure*}

\lev{0}~\textbf{Binary accuracy} provides a directional signal but little diagnostic information; on a balanced set, a reject-everything system and a random one are indistinguishable.
\lev{1}~\textbf{Concern detection} measures \emph{concern recall} (the fraction of official concerns with a strict match) and \emph{phantom rate} (the fraction of agentic concerns with no strict match); both are verdict-blind. Not all phantoms are equal: many are legitimate concerns the official reviewers did not raise, while others are fabricated or over-severe. We distinguish \emph{harmful phantoms} (fatal/major unmatched concerns on accepted papers) from benign ones at Level~3.
\lev{2}~\textbf{Verdict-stratified metrics} split Level~0--1 metrics by accepted versus rejected papers, revealing reject-heavy behavior that overall accuracy hides.
\lev{3}~\textbf{Decision-aware metrics} introduce the \emph{false decisive rate} (FDR): among agentic concerns on accepted papers, what fraction is marked decisive?
Under our AC-aligned operationalization, the count of decisive blockers on accepted papers is zero.\footnote{See the assumption discussion in \S\ref{sec:matchgraph} and \S\ref{sec:discussion}.
Any non-zero FDR on accepted papers represents concerns the AC did not treat
as blocking under this operationalization.}
On rejected papers, where real decisive blockers exist, two complementary metrics replace FDR: \emph{decisive precision (strict)} measures the fraction of agentic decisive flags that match an official decisive blocker, and \emph{phantom decisive rate} measures the fraction of total output that consists of fabricated decisive concerns (unmatched decisive flags).
\emph{Resolved-escalation rate} measures how often the system re-escalates concerns the AC marked resolved when the corresponding fix is visible in the reviewed PDF.
We also decompose concerns into \emph{decision-relevant} (correct matches at appropriate severity on rejected papers, or constructive feedback on accepted ones) and \emph{decision-harmful} (re-escalation of dismissed or resolved concerns, harmful phantoms on accepted papers, and severity under-rating or missed blockers on rejected papers), reporting both components separately rather than collapsing them.
\lev{4}~\textbf{Rebuttal-aware decomposition} computes recall separately for each AC treatment category (decisive blocker, unresolved, accepted limitation, resolved), revealing the system's \emph{attention profile}.

All Level~3--4 metrics can also be computed at \emph{top-$K$} by
restricting attention to the $K$ most severe agentic concerns per paper.
This is not a replacement for full-review metrics; it answers a different
usage question. Full-review metrics evaluate the complete output. Top-$K$
metrics evaluate the prioritized core that a reader is most likely to act
on. They are especially informative when systems generate 2--3$\times$
different concern volumes: a low full-review FDR can reflect genuine
calibration or simple concern dilution. That distinction matters because
researchers triaging a review will typically focus on the top few concerns
first (\S\ref{sec:discrimination}).

\subsection{Measurement Validation}
\label{sec:validation}

An independent auditor using GPT-5.4 Pro re-verified 191 candidate
edges across 9 held-out papers. The auditor worked from local edge
worksheets rather than downstream system scores. Two independent runs of the auditor
on the same edges agreed on 96.9\% of labels; Cohen's $\kappa$ was 0.918 for
verdict and 0.946 for both match type and severity, indicating near-perfect
consistency (Appendix~\ref{app:validation_details}). In focused extraction audits, 16
of 18 official concern sheets and 51 of 54 agentic sheets were rated
satisfactory on completeness and label quality; a broader completeness audit covered all 48 official concern sheets in the full benchmark, with no
extracted concern unsupported by the source text. The dominant matching error
is \emph{scope inflation} (where one concern bundles the other's complaint with additional independent demands; 59\% of errors), not false matches. A phantom-quality audit
($N{=}200$, independently rated by two annotators, 19 inter-annotator disagreements resolved by human adjudication) further
suggests that many phantoms are legitimate concerns the official reviewers did not raise.

Because match graph construction uses LLMs to evaluate LLM-generated reviews,
circularity remains a methodological risk. We mitigate it through cross-model
validation with a different model family ($\kappa > 0.91$), human
adjudication of all auditor disagreements and a random sample of agreement
cases, 32 calibration exemplars spanning 6 error categories
(Appendix~\ref{app:exemplars}), and extraction audits of all 48 official
concern sheets plus a stratified sample of agentic sheets for severity
accuracy and AC-treatment coding (\S\ref{sec:discussion};
Appendix~\ref{app:validation_details}).

\section{Pilot Study Setup}
\label{sec:setup}

We demonstrate the framework on four public AI review systems applied to 48 papers from ICLR~2026, NeurIPS~2025, and ICML~2025, all in AI safety/alignment (24 accepted, 24 rejected; 670 official concerns, 79 decisive blockers).
The goal is to show what concern-level diagnostics reveal about reviewer behavior, not to produce a leaderboard.
The sample is sufficient for a framework demonstration but not for population-level estimates; we state that limitation explicitly in \S\ref{sec:discussion}.

\paragraph{Data curation.}
Papers were sourced from OpenReview and filtered to the AI safety/alignment domain to control for topic-specific reviewing norms.
Selection prioritized review quality: each paper has $\geq$3 substantive reviews, an unambiguous AC decision with articulated reasoning, and $\geq$2 extractable technical concerns.
PDFs were sanitized to remove decision-revealing metadata.
We also tracked which paper version each system reviewed and whether fixes to resolved concerns were visible in that version.
Of the 170 concerns marked resolved by the AC, 40 (24\%) had fixes absent from the reviewed PDF; these remain valid detection targets rather than calibration failures.
Details appear in Appendix~\ref{app:curation}.
Systems receive the camera-ready PDF for accepted papers and the original submission for rejected papers; concern extraction reads the full OpenReview record including rebuttals and meta-reviews.

We evaluate six configurations (Table~\ref{tab:systems}):
\textbf{System~L}~\citep{liang2024can} (single-prompt zero-shot),
\textbf{System~A}~\citep{lu2024aiscientist} (iterative reflection),
\textbf{System~O}~\citep{oar2024} (progressive structured review),
and \textbf{System~M}~\citep{darcy2024marg} (multi-agent swarm).
Systems~L, A, and~O run on Claude Opus to control for model effects; L~and~A additionally run on GPT-4o.
System~M runs on GPT-4o only using its native OpenAI API (Opus runs produced degenerate output (repetitive or truncated reviews) under our SDK adaptation); its metrics reflect combined method and model effects (\S\ref{sec:model}).
Each configuration is run 3 times. All systems use published code with adaptations documented in Appendix~\ref{app:implementation}. Runs were conducted between February and April 2026 using Claude Opus (Anthropic API) and GPT-4o (OpenAI API) as named; exact API snapshot identifiers for each configuration are archived with the release artifacts.

\begin{table}[t]
\small
\begin{minipage}[t]{0.45\columnwidth}
\centering
\caption{Systems analyzed. Concerns/paper is the average number of atomic, deduplicated concerns per paper (3-run mean across all 48 papers). GPT-4o single-agent systems produce roughly half as many concerns as their Opus counterparts.}
\label{tab:systems}
\scriptsize
\begin{tabular}{lcc}
\toprule
\textbf{System} & \textbf{Model} & \textbf{Concerns/paper} \\
\midrule
System~L & Opus & 10.6 \\
System~A & Opus & 11.5 \\
System~O & Opus & 8.3 \\
System~L & GPT-4o & 5.2 \\
System~A & GPT-4o & 4.8 \\
System~M & GPT-4o & 10.1 \\
\bottomrule
\end{tabular}
\end{minipage}\hfill
\begin{minipage}[t]{0.52\columnwidth}
\centering
\caption{Detection and calibration can point in different directions. Across the Opus systems, the configuration with the strongest rejected-paper recall still shows weak accepted-paper calibration. Accepted acc.\ = accepted-paper verdict accuracy (pipeline inference; see Appendix~\ref{app:verdict_audit} for sensitivity); false decisive rate and resolved-escalation are computed on accepted papers.}
\label{tab:naive_vs_deep}
\scriptsize
\begin{tabular}{lccc}
\toprule
\textbf{Metric (Level)} & \textbf{L} & \textbf{A} & \textbf{O} \\
\midrule
\multicolumn{4}{l}{\emph{Detection (rejected)}} \\
Recall (L1) & {.44} & .44 & .17 \\
\midrule
\multicolumn{4}{l}{\emph{Calibration (accepted)}} \\
Accepted acc.\ (L2) & .028 & .083 & .347 \\
False dec.\ rate (L3) & .49 & {.36} & .37 \\
Resolved-esc.\ (L3) & .63 & {.60} & .62 \\
\bottomrule
\end{tabular}
\end{minipage}
\end{table}

\paragraph{Severity extraction.}
\label{sec:extraction}
Baselines expose severity very differently: System~A provides structured weaknesses with numeric scores and an explicit decision; System~L has a ``reasons for rejection'' section; System~M labels concerns as critical/major/minor; System~O provides no native severity labels.
We therefore normalize severity and decisive flags with an extraction pass over the full review text, using structural cues, language intensity, and available scores.
In human audits of 54 paper-method pairs, 51 extraction sheets were rated satisfactory for completeness and support, and none contained unsupported extracted concerns (\S\ref{sec:validation}).
For Systems~L, A, and~O, the resulting severity labels are our normalized interpretation rather than native system outputs; System~M is closest to native severity (Appendix~\ref{app:severity_examples}).
Absolute FDR values should therefore be read as approximate, but the paper's main claims rest on replicated qualitative patterns such as flat severity profiles across accepted and rejected papers.
Level~3--4 metrics thus evaluate the interaction between review generation and severity interpretation, not only the underlying generator (\S\ref{sec:discussion}).

\paragraph{Verdict extraction.}
Evaluating verdict accuracy requires an accept/reject label for each review, but most systems do not emit one explicitly; their reviews carry signals of acceptance or rejection without a binary field.
We therefore apply a verdict inference procedure that maps each free-form review to an accept/reject label.
Both System~A configurations produce an explicit \texttt{Decision} field; the other four configurations do not.
System~L organizes output into ``reasons for acceptance'' and ``reasons for rejection'' sections, from which a verdict is inferred by the extraction pipeline.
Systems~O and~M produce no native verdict or scores; we infer the verdict from overall review tone and the presence or absence of blocking-level language (see Appendix~\ref{app:severity_examples} for worked examples of each system's output structure).
The extraction pipeline uses a default-REJECT rule for ambiguous cases, which may inflate the reject rate for systems that write analytically balanced reviews.
Because verdict accuracy numbers for Systems~L, O, and~M reflect this inference procedure rather than native system decisions, we conducted a verdict inference audit with two independent raters and human adjudication (Appendix~\ref{app:verdict_audit}).
The audit shows that verdict-level findings are sensitive to the inference method, while the paper's concern-level diagnostics---recall, FDR, decisive precision, phantom rates, attention profiles, ICC, and top-$K$ analyses, all computed or stratified by official verdict---are unchanged by how accept/reject is inferred.
We report pipeline-inferred verdicts throughout as the primary analysis, with sensitivity ranges from the audit noted where relevant.

\section{What Concern Alignment Reveals}
\label{sec:findings}

Each subsection presents a diagnostic finding invisible to the level below. Table~\ref{tab:naive_vs_deep} previews the core tension: at \lev{1}, System~L~(Opus) appears to be the best Opus system (highest recall); by \lev{3}, its calibration failures are exposed.

\subsection{Binary Accuracy Masks Structural Differences}
\label{sec:binary}

Most systems do not produce explicit accept/reject decisions (\S\ref{sec:setup}); we infer verdicts from review tone using a pipeline with a default-REJECT rule.
Verdict-stratified accuracy under this pipeline (Appendix~\ref{app:verdict_split}, Table~\ref{tab:verdict_split}) shows that several configurations exhibit reject-heavy profiles, and that identical overall accuracy can conceal structurally different behaviors.
A verdict inference audit with two independent methods and two independent raters (Appendix~\ref{app:verdict_audit}, Table~\ref{tab:verdict_sensitivity}) confirms that these numbers are sensitive to the inference method: the model-effect swing ranges from 46 to 96 percentage points depending on who reads the review and what rules they apply.
The sensitivity of verdict accuracy to the inference method is itself diagnostic: it confirms that verdict-level evaluation provides an unreliable measurement surface for systems not designed to produce explicit recommendations.
The concern-level diagnostics used in the remainder of this section---recall, FDR, decisive precision, phantom rates, attention profiles, ICC, and top-$K$ analyses---are invariant to the verdict inference method because they are computed or stratified by official verdict rather than predicted verdict.

\begin{table}[t]
\centering
\small
\caption{Core concern-level metrics (3-run mean $\pm$ std). Rcl = concern recall on rejected papers; Dec.\ rcl = decisive-blocker recall on rejected papers; FDR = false decisive rate on accepted papers; Res.\ esc = resolved-escalation on accepted papers; DecP = decisive precision (strict) on rejected papers; PhDec = phantom decisive rate on rejected papers. $^\dagger$A~(4o) has \emph{higher} FDR despite fewer concerns.}
\label{tab:core_metrics}
\scriptsize
\setlength{\tabcolsep}{3pt}
\begin{tabular}{lcccccc}
\toprule
\textbf{Sys.} & \textbf{Rcl} & \textbf{Dec.\ rcl} & \textbf{FDR} & \textbf{Res.\ esc} & \textbf{DecP} & \textbf{PhDec} \\
& \scriptsize{(rej)} & \scriptsize{(rej)} & \scriptsize{(acc)} & \scriptsize{(acc)} & \scriptsize{(rej)} & \scriptsize{(rej)} \\
\midrule
L (Op) & .44{\tiny$\pm$.02} & .68{\tiny$\pm$.01} & .49{\tiny$\pm$.03} & .63{\tiny$\pm$.04} & .33{\tiny$\pm$.02} & .14{\tiny$\pm$.03} \\
A (Op) & .44{\tiny$\pm$.01} & .65{\tiny$\pm$.01} & .36{\tiny$\pm$.02} & .60{\tiny$\pm$.05} & .36{\tiny$\pm$.02} & .11{\tiny$\pm$.02} \\
O (Op) & .17{\tiny$\pm$.01} & .22{\tiny$\pm$.04} & .37{\tiny$\pm$.03} & .62{\tiny$\pm$.08} & .17{\tiny$\pm$.01} & .26{\tiny$\pm$.03} \\
L (4o) & .25{\tiny$\pm$.01} & .26{\tiny$\pm$.00} & .25{\tiny$\pm$.02} & .61{\tiny$\pm$.07} & .32{\tiny$\pm$.07} & .08{\tiny$\pm$.02} \\
A (4o)$^\dagger$ & .22{\tiny$\pm$.02} & .37{\tiny$\pm$.07} & .55{\tiny$\pm$.02} & .70{\tiny$\pm$.11} & .36{\tiny$\pm$.04} & .18{\tiny$\pm$.04} \\
M (4o) & .27{\tiny$\pm$.04} & .31{\tiny$\pm$.03} & .10{\tiny$\pm$.06} & .34{\tiny$\pm$.06} & .18{\tiny$\pm$.10} & .09{\tiny$\pm$.04} \\
\bottomrule
\end{tabular}
\end{table}

Concern-level evaluation also enables a stability diagnostic that verdict-level analysis cannot: the intraclass correlation coefficient (ICC), which measures whether the same paper receives similar scores across independent runs.
On rejected papers, concern recall ICC(2,1) (a two-way random-effects model treating papers as subjects and runs as raters) exceeds verdict ICC for 5 of 6 systems (Appendix~\ref{app:icc}).\footnote{ICC on binary verdicts has known limitations: with near-ceiling or near-floor proportions, ICC can be near zero despite high raw agreement. We report it to enable the comparison with concern recall ICC, but note that Fleiss' $\kappa$ would be an alternative for the binary case.}
For near-universal reject profiles under the pipeline (L~(Opus), A~(GPT-4o)), verdict ICC is near zero, yet concern recall ICC remains meaningful (0.69 and 0.41).
Even when the verdict carries little per-paper information, concern-level analysis still extracts diagnostic signal (Appendix~\ref{app:icc}).

On Paper~D, all three Opus systems correctly reject (100\% binary accuracy), yet System~A catches all 3 AC content-related decisive blockers while System~O catches none, focusing only on notation (a fourth blocker, concerning reviewer engagement during discussion, is excluded as invisible to PDF-only systems). Concern alignment reveals a quality gap that binary accuracy misses (Appendix~\ref{app:casestudies}); throughout the case-study appendix we refer to papers by letter designation and keep identities for supplementary release.

\subsection{Detection Without Discrimination}
\label{sec:discrimination}

\begin{wrapfigure}{r}{0.44\columnwidth}
\vspace{-12pt}
\centering
\includegraphics[width=0.38\columnwidth]{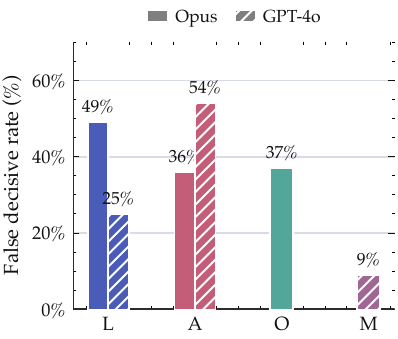}
\caption{False decisive rate on accepted papers. Most single-agent systems mark 25--55\% of concerns decisive under the AC-aligned operationalization.}
\label{fig:fdr_bar}
\vspace{-10pt}
\end{wrapfigure}

Table~\ref{tab:core_metrics} and Figure~\ref{fig:fdr_bar} show the core calibration pattern.
FDR on accepted papers ranges from 0.10 (System~M) to 0.55 (System~A~(GPT-4o)).
Most single-agent systems show essentially flat decisive behavior across verdicts: System~L~(Opus), for example, marks 49\% of concerns as decisive on accepted papers and 52\% on rejected ones.
Under our AC-aligned operationalization, the accepted-paper decisive-blocker count is zero, so these systems are not only over-escalating; they are failing to modulate decision weight with paper quality.
System~M is the exception (10\% accepted, 21\% rejected), though part of that advantage reflects a low-decisive-flag profile rather than uniformly strong calibration.
A long list can recover some official concerns through volume rather than selective attention.
This is the volume-without-discrimination pattern: coverage that arises from
sheer concern count rather than selective attention.

On rejected papers, where real decisive blockers exist, decisive precision (strict) and phantom decisive rate decompose the decisive signal (Table~\ref{tab:core_metrics}).
Decisive precision ranges from 0.17 (System~O) to 0.36 (System~A~(Opus)), meaning at best only 36\% of decisive flags identify a concern the AC also treated as a decisive blocker.
Phantom decisive rate ranges from 0.08 (System~L~(GPT-4o)) to 0.26 (System~O): the fraction of total output that consists of fabricated decisive concerns.
Systems~L and~A~(Opus) achieve the best combination (DecPrec 0.33--0.36, PhDecRate 0.11--0.14), while System~O produces the most noise (61\% of its decisive flags are phantoms).
System~M shows low phantom volume (PhDecRate 0.09) but also low precision (0.18), consistent with its conservative decisive threshold that avoids false alarms at the cost of missing real blockers.

The calibration failure is visible at the level of individual concerns.
On the accepted-paper example in Figure~\ref{fig:matchgraph}, the official review notes that the core evaluation metric ``lacks formal validation of its estimator'' (\emph{moderate, resolved}); System~L~(Opus) frames the same issue as the estimator ``does not approximate the theoretical quantity it claims to measure'' (\emph{fatal, decisive}).
The concern family is similar, but the severity and decision weight are not.
Concern recall rewards the match; FDR penalizes the escalation.

Top-$K$ analysis (Appendix~\ref{app:topk}) reveals a second pattern: some low-FDR systems achieve their aggregate rate partly through \emph{concern dilution} (producing many low-severity concerns that reduce the decisive fraction without improving calibration) rather than selective prioritization.
System~M's FDR rises from 0.10 to 0.28 at $K{=}3$. Systems~L and~A~(Opus) show the opposite pattern: at $K{=}3$, 90\% or more of top concerns are marked decisive (FDR $\geq$ 0.90). 
Full $K$-curves and exact values appear in Appendix~\ref{app:topk}.

Top-$K$ is useful because reviews are consumed as prioritized concern lists.
A researcher triaging feedback will typically focus on the top few concerns first. Full-review and top-$K$ metrics therefore answer complementary questions: what the system says in total, and what it tells the reader to focus on first.

\subsection{Model vs.\ Method Effects}
\label{sec:model}

Table~\ref{tab:model_method} presents Systems~L and~A on both Opus and GPT-4o.
These paired comparisons do not estimate the relative size of model and method effects, but they make a known concern concrete: model choice alone materially shifts both verdict bias and concern calibration, and the framework quantifies which diagnostic dimensions change.
Under the pipeline, System~L flips from reject-heavy on Opus to majority-accept on GPT-4o; across alternative inference methods, the accepted-paper-accuracy swing ranges from 46 to 96 percentage points (Table~\ref{tab:verdict_sensitivity}).\footnote{Three inference methods (pipeline with default-REJECT, independent tone reading, concern-gate-based rules) applied by two independent raters ($\kappa \geq 0.74$), plus human adjudication of 54 disagreement cases. One rater's tone reading assigns L~(Opus) 42\% accepted-paper accuracy, an outlier against all other method/rater combinations ($\leq$\,4\%); see Appendix~\ref{app:verdict_audit}.}

\begin{table}[t]
\small
\begin{minipage}[t]{0.48\columnwidth}
\centering
\caption{Model choice with fixed prompts produces large shifts. A~(GPT-4o) has \emph{higher} false decisive rate despite fewer concerns; in the column headers, ``4o'' abbreviates GPT-4o. Acc-corr reflects pipeline inference; see Appendix~\ref{app:verdict_audit} for sensitivity.}
\label{tab:model_method}
\scriptsize
\begin{tabular}{lcccc}
\toprule
\textbf{Metric} & \textbf{L\,Opus} & \textbf{L\,4o} & \textbf{A\,Opus} & \textbf{A\,4o} \\
\midrule
Conc./paper & 10.6 & 5.2 & 11.5 & 4.8 \\
Acc-corr. & 2.8\% & 63.9\% & 8.3\% & 4.2\% \\
Recall (rej) & .44 & .25 & .44 & .22 \\
Dec.\ rcl (rej) & .68 & .26 & .65 & .37 \\
FDR (acc) & .49 & .25 & .36 & {.55} \\
\bottomrule
\end{tabular}
\end{minipage}\hfill
\begin{minipage}[t]{0.48\columnwidth}
\centering
\caption{Recall by AC treatment on rejected papers (L4). Dec.\ = decisive blocker, Unres.\ = unresolved, Res.\ = resolved, and Gap is decisive-blocker recall minus resolved-concern recall. Negative gaps (\textbf{bold}) indicate inverted attention.}
\label{tab:attention}
\scriptsize
\begin{tabular}{lcccc}
\toprule
\textbf{System} & \textbf{Dec.} & \textbf{Unres.} & \textbf{Res.} & \textbf{Gap} \\
\midrule
L (Opus) & .68 & .47 & .48 & +20pp \\
A (Opus) & .66 & .44 & .49 & +17pp \\
O (Opus) & .18 & .20 & .20 & {$-$2pp} \\
L (GPT-4o) & .24 & .29 & .33 & {$-$9pp} \\
A (GPT-4o) & .37 & .34 & .25 & +12pp \\
M (GPT-4o) & .29 & .33 & .38 & {$-$9pp} \\
\bottomrule
\end{tabular}
\end{minipage}
\end{table}

The effect is not uniform: System~A~(GPT-4o) stays at 4.2\% accepted accuracy while achieving \emph{higher} FDR (0.55) than its Opus counterpart (0.36), despite generating half as many concerns.
On Paper~H (rejected), System~L~(Opus) catches all three AC decisive blockers in every run; System~L~(GPT-4o) misses all three and misdiagnoses a substance problem as a clarity problem (Appendix~\ref{app:casestudies}).
Published comparisons that use different models per system therefore interleave method and model effects; concern alignment makes that entanglement visible and quantifiable.

\subsection{Attention Profiles and Decision-Weight Calibration}
\label{sec:attention}
\begin{wrapfigure}{r}{0.44\columnwidth}
\vspace{-12pt}
\centering
\includegraphics[width=0.38\columnwidth]{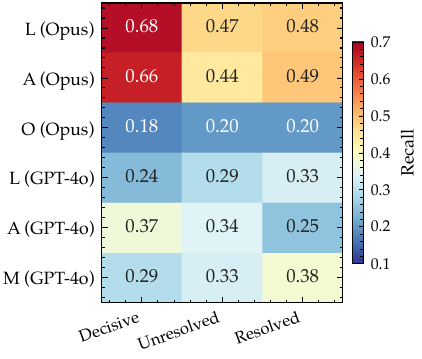}
\caption{Recall by AC treatment on rejected papers. Positive gaps indicate greater attention to decisive blockers than to resolved concerns.}
\label{fig:attention_heatmap}
\vspace{-10pt}
\end{wrapfigure}
Level~4 decomposes recall by AC treatment (Table~\ref{tab:attention}), asking whether systems preferentially recover the concerns that remained
consequential after discussion and rebuttal rather than the ones the AC
treated as resolved or non-blocking.
Systems~L and~A~(Opus) show positive gaps: recall on decisive blockers exceeds recall on resolved concerns by 17--20 percentage points.
Systems~O, L~(GPT-4o), and~M show flat or inverted profiles, meaning they recover resolved concerns at least as readily as decisive blockers (Figure~\ref{fig:attention_heatmap}).

Overall recall hides this structure: System~L~(Opus) has 0.44 overall recall, but 0.68 on decisive blockers versus 0.48 on resolved concerns.
The gap varies from +20pp to $-$9pp across systems.
This over-escalation appears concretely: on Paper~B, an official reviewer hedges that scaffolding ``may be unfamiliar to some models'' (marked \texttt{resolved} after rebuttal); System~L~(Opus) rephrases the same issue as a \textbf{fundamental} confound and treats it as decisive.
The issue is detected, but its decision weight is inflated.
Resolved-escalation rate measures that failure directly.

\subsection{Phantoms and Decision Impact}
\label{sec:phantoms}
\label{sec:netneg}

Phantom rates on accepted papers range from 42\% to 83\%. A dual-annotated audit ($N{=}200$) suggests many phantoms are legitimate concerns the official reviewers did not raise, though transfer from that audit system to Systems~L/A/O/M is only suggestive. On Paper~C (accepted spotlight), System~O raises a \textbf{fatal} phantom with a fabricated counterexample that no official reviewer or AC identified as an error in the paper, whereas System~L's phantoms on the same paper are mostly extra but plausible feedback. \emph{Harmful phantom rate} (fatal/major phantoms on accepted papers) separates damaging hallucinations and over-escalations from benign extra feedback.

We therefore decompose output into decision-relevant and decision-harmful components instead of reporting a single net score. On rejected papers, harmful components exceed relevant ones across configurations (derived from the match graphs underlying Appendix~\ref{app:detailed}). On accepted papers the mix is more heterogeneous, which makes the decomposition useful: it separates harmful phantoms, resolved-concern re-escalation, and missed decisive issues.

Paper~F illustrates the point: System~A achieves 64\% recall with 25\% phantom rate yet still rejects the paper because it escalates a minor presentation issue and downrates a major methodological one.

\section{Related Work}
\label{sec:related}

Recent work on AI peer review spans direct prompting, iterative reflection, structured review pipelines, multi-agent reviewers, fine-tuned reviewer models, and pairwise-comparison reformulations \citep{liang2024can,lu2024aiscientist,darcy2024marg,oar2024,gao2025reviewagents,idahl2025openreviewer,zou2026diagpaper,zhang2025pairwise}. The closest evaluation papers compare facet distributions \citep{shin2025blindspots}, score holistic review quality \citep{garg2025revieweval}, reconstruct premises to test whether review claims are misinformed \citep{ryu2025reviewscore}, or benchmark weakness and limitation discovery \citep{xu2025limitgen,lou2025aaar}. Concern alignment instead evaluates free-form concern instances, models both misses and phantoms, measures decision weight as well as detection, and grounds evaluation in post-rebuttal AC treatment; Appendix~\ref{app:related_extended} gives the extended discussion.

\section{Discussion and Limitations}
\label{sec:discussion}

\paragraph{Scope.} Our pilot covers 48 papers in one domain and four systems. That is enough to expose failure modes hidden by verdict accuracy, but not enough for population-level claims or fine-grained rankings.

\paragraph{Operational anchor.} The framework evaluates review quality as feedback that helps researchers improve papers, not as a way to optimize for conference outcomes. We use AC decisions as a noisy post-deliberation anchor for which concerns were treated as consequential. That gives the metrics a concrete reference point, but it does not imply that accepted papers are free of serious weaknesses or that rejected papers were clearly below the bar \citep{cortes2021inconsistency}. False decisive rate on accepted papers is therefore defined relative to the official review rationale.
Concern-level analysis is specifically designed to be more robust to this noise than verdict agreement: by decomposing evaluation into inspectable concern units, many of which are independently verifiable regardless of the final decision, the framework extracts diagnostic signal even when verdicts are unreliable.
The structural patterns it identifies (flat severity profiles, inverted attention, concern dilution) are robust across 48 papers in a way that individual verdict disagreements are not.

\paragraph{Verdict inference.} Most systems we evaluated do not produce native accept/reject decisions; we infer verdicts from review tone using a separate extraction pipeline (\S\ref{sec:setup}).
A two-rater audit with human adjudication (Appendix~\ref{app:verdict_audit}) shows that verdict-level findings are sensitive to the inference method, whereas the concern-level diagnostics used throughout the paper---recall, FDR, decisive precision, phantom rates, attention profiles, ICC, and top-$K$ analyses---are unaffected because they are computed or stratified by official verdict rather than predicted verdict.
Across audited methods, the L~Opus $\to$ L~GPT-4o accepted-paper-accuracy swing ranges from 46 to 96 percentage points, with the pipeline estimate lying inside that range (Table~\ref{tab:verdict_sensitivity}).
When systems do not emit explicit recommendations, verdict agreement becomes an unstable measurement target.
Separately, this pilot suggests that verdict readability and severity calibration can co-vary, but the relationship is not monotone across the six configurations.
Some hard-to-read reviews also show flatter severity profiles, but other configurations break that pattern.
We therefore treat verdict ambiguity and calibration as related but distinct observations, not as evidence of a general association or mechanism.

\paragraph{Measurement validity.} Match graphs are constructed with LLM assistance, so circularity remains a methodological risk. Cross-model validation with GPT-5.4 Pro against a Claude-based primary annotator, human adjudication, and extraction audits reduce that risk but do not remove shared biases. For Systems~L, A, and~O, severity is inferred by our extraction pipeline rather than emitted natively, so Level~3--4 metrics partly evaluate that inference step. We also avoid penalizing a concern as re-escalated when discussion resolved it but the fix is not visible in the PDF the system reviewed. System~M was run only on GPT-4o, so its profile remains a combined model-plus-method effect; with only 3 runs, we emphasize replicated qualitative patterns over fine point estimates.

\section{Conclusion}
\label{sec:conclusion}

Binary accuracy cannot tell us \emph{how} a system fails or \emph{what to fix}. Concern alignment does so by evaluating the unit of review that authors and readers actually act on.

Across the pilot, each level of the ladder exposed a different failure mode: verdict stratification revealed reject-heavy behavior; decision-aware metrics quantified severity miscalibration; rebuttal-aware recall exposed inverted attention; and top-$K$ analysis separated calibrated prioritization from concern dilution.

The clearest lesson is that finding issues is not enough. The systems we examined often detect real concerns yet still misjudge their decision weight, especially on accepted papers. Concern-level diagnostics may therefore be useful beyond peer review for evaluative AI systems that generate critiques, due-diligence reports, or risk assessments. We view concern alignment as an evaluation substrate: a way to audit which concerns systems identify, how they weight them, and whether those priorities align with the review rationale that informed the final assessment.

\section*{Acknowledgements}
The work of M. Jin was supported in part by the National Science Foundation (NSF) under grants ECCS-2500368, ECCS-2331775, and IIS-2312794, the Commonwealth Cyber Initiative, and the Amazon--Virginia Tech Initiative for Efficient and Robust Machine Learning.

\paragraph{Code and data availability.}
The concern-alignment framework, evaluation pipeline, and aggregate benchmark metrics are available at \url{https://github.com/jinming99/reviewer-under-review} and \url{https://jinming99.github.io/reviewer-under-review/}. The public release also includes a 9-paper illustrative slice with linked raw artifacts; this slice is distinct from the anonymized case studies referenced in the paper's appendix, and the 48-paper pilot papers themselves are not redistributed.

\bibliographystyle{colm2026_conference}
\bibliography{paper}

\newpage
\appendix

\section{Detailed Metric Definitions}
\label{app:metrics}

We restate the notation here for standalone readability.
For a paper $p$, let $\mathcal{O}_p = \{o_1,\ldots,o_m\}$ denote the set of official concerns and $\mathcal{A}_p = \{a_1,\ldots,a_n\}$ the set of agentic concerns.
Let $E_{\text{strict}}(p) \subseteq \mathcal{O}_p \times \mathcal{A}_p$ be the set of \texttt{exact} and \texttt{partial} matches only.
Let $v_p$ and $\hat{v}_p$ denote the official and agentic verdicts.
Let $t(o_i)$ denote the AC-treatment label of official concern $o_i$, let $s(a_j)$ denote the severity of agentic concern $a_j$, and let $\mathrm{dec}(a_j) \in \{0,1\}$ denote whether $a_j$ is marked decisive.
When metrics are stratified by decision, $\mathcal{P}_{\text{acc}}$ and $\mathcal{P}_{\text{rej}}$ denote the accepted and rejected paper subsets, and $\mathcal{A}_{\text{acc}}=\bigcup_{p\in\mathcal{P}_{\text{acc}}}\mathcal{A}_p$ denotes the multiset of all agentic concerns on accepted papers.

\paragraph{Level~0: Binary accuracy.}
\begin{equation}
\text{Acc} = \frac{1}{N} \sum_{p=1}^{N} \mathbf{1}[\hat{v}_p = v_p]
\end{equation}

\paragraph{Level~1: Concern recall (strict).}
For a single paper,
\begin{equation}
\text{Recall}(p) = \frac{|\{o_i \in \mathcal{O}_p : \exists\, e_{ij} \in E_{\text{strict}}(p)\}|}{|\mathcal{O}_p|}
\end{equation}
Reported system-level recall is the mean of $\text{Recall}(p)$ over the relevant paper subset.

\paragraph{Level~1: Phantom rate.}
For a single paper,
\begin{equation}
\text{Phantom}(p) = \frac{|\{a_j \in \mathcal{A}_p : \nexists\, e_{ij} \in E_{\text{strict}}(p)\}|}{|\mathcal{A}_p|}
\end{equation}
Reported system-level phantom rate is again the mean over papers.

\paragraph{Level~2: Verdict-stratified metrics.}
All Level~0--1 metrics can be computed separately on $\mathcal{P}_{\text{acc}}$ and $\mathcal{P}_{\text{rej}}$.

\paragraph{Level~3: False decisive rate.}
\begin{equation}
\text{FDR}_{\text{acc}} = \frac{\sum_{a_j \in \mathcal{A}_{\text{acc}}} \mathrm{dec}(a_j) - |\mathcal{E}|}{|\mathcal{A}_{\text{acc}}|}
\end{equation}
That is, among all agentic concerns raised on accepted papers, what fraction carry a decisive flag after excusing flags that correctly detect resolved-but-unfixed issues?
Here $\mathcal{E}$ is the set of decisive flags excused because they match a resolved official concern whose fix is absent from the reviewed PDF (correct detection, not a false alarm).

\paragraph{Level~3: Decisive precision and phantom decisive rate (rejected papers).}
On rejected papers, some decisive flags correctly identify real blockers, so FDR is not meaningful. Two metrics replace it:
\begin{equation}
\text{DecPrec}_{\text{strict}} = \frac{|\{a_j \in \mathcal{A}_{\text{rej}} : \mathrm{dec}(a_j)=1,\; \exists\, e_{ij} \in E_{\text{strict}} \text{ with } t(o_i)=\texttt{decisive\_blocker}\}|}{|\{a_j \in \mathcal{A}_{\text{rej}} : \mathrm{dec}(a_j)=1\}|}
\end{equation}
That is, among agentic concerns flagged decisive on rejected papers, what fraction matches an official concern that the AC also treated as a decisive blocker?
\begin{equation}
\text{PhDecRate} = \frac{|\{a_j \in \mathcal{A}_{\text{rej}} : \mathrm{dec}(a_j)=1,\; \nexists\, e_{ij} \in E_{\text{strict}}\}|}{|\mathcal{A}_{\text{rej}}|}
\end{equation}
That is, what fraction of total agentic output on rejected papers consists of decisive flags with no official match (fabricated blockers)?

\paragraph{Level~3: Resolved-escalation rate.}
Let $\mathrm{pdf}(o_i) \in \{0,1\}$ indicate whether a concern marked resolved by the AC is actually addressed in the reviewed PDF (\texttt{addressed\_in\_pdf}).
We restrict to concerns with $t(o_i)=\texttt{resolved}$ and $\mathrm{pdf}(o_i)=1$; concerns resolved only in rebuttal text remain valid detection targets rather than calibration failures.
\begin{equation}
\text{ResEsc} =
\frac{|\{(o_i,a_j)\in E_{\text{strict}}(p) : t(o_i)=\texttt{resolved},\; \mathrm{pdf}(o_i)=1,\; s(a_j)\in\{\texttt{fatal},\texttt{major}\}\}|}
{|\{(o_i,a_j)\in E_{\text{strict}}(p) : t(o_i)=\texttt{resolved},\; \mathrm{pdf}(o_i)=1\}|}
\end{equation}

\paragraph{Level~3: Decision-relevant and decision-harmful decomposition.}
Let $R_p$ be the set of decision-relevant agentic concerns for paper $p$ (strict matches to official concerns at appropriate severity on rejected papers, or constructive non-blocking feedback on accepted papers), and let $H_p$ be the set of decision-harmful concerns.
On accepted papers, harmful concerns include re-escalation of dismissed or resolved concerns (the latter only when the fix appears in the reviewed PDF) to fatal/major severity, and harmful phantoms (unmatched fatal/major concerns).
On rejected papers, harmful concerns include severity under-rating of fatal/major official concerns to minor/moderate and missed decisive blockers.
We report the decision-relevant rate ($|R_p|/|\mathcal{A}_p|$) and decision-harmful rate ($|H_p|/|\mathcal{A}_p|$) separately because the components reveal \emph{where} harm originates.

\paragraph{Level~4: Recall by AC treatment.}
For each treatment category $t \in T$ (where $T$ is the set of AC-treatment categories) and paper $p$,
\begin{equation}
\text{Recall}_t(p) = \frac{|\{o_i \in \mathcal{O}_p : t(o_i)=t,\; \exists\, e_{ij} \in E_{\text{strict}}(p)\}|}{|\{o_i \in \mathcal{O}_p : t(o_i)=t\}|}
\end{equation}
Reported treatment-specific recall is the mean of $\text{Recall}_t(p)$ over the relevant paper subset.

\paragraph{Volume-normalized variants (top-$K$).}
Any Level~3--4 metric $M$ can be computed at top-$K$ by restricting to the $K$ most severe agentic concerns per paper (ranked by severity, with ties broken by decisive flag):
\begin{equation}
M@K = M\big(\mathcal{A}^{(K)}_p\big) \quad\text{where}\quad \mathcal{A}^{(K)}_p = \text{top-}K(\mathcal{A}_p, \text{severity})
\end{equation}
This normalizes for review length: systems generating 4.8 to 11.5 concerns per paper are compared at equal concern budget. FDR@$K$ is especially diagnostic because it tests whether a system's \emph{most severe} concerns are calibrated, not just whether the overall proportion of decisive flags is low.

\section{Per-System Detailed Metrics}
\label{app:detailed}

Table~\ref{tab:full_metrics} presents the complete set of verdict-stratified metrics for all system configurations.

\begin{table*}[t]
\centering
\small
\caption{Full verdict-stratified metrics (3-run means). Acc = accepted papers, Rej = rejected papers. DecPrec = decisive precision (strict), PhDec = phantom decisive rate (both on rejected papers only). Per-run bootstrap 95\% CIs (10,000 paper-level resamples) available in supplementary materials.}
\label{tab:full_metrics}
\begin{tabular}{lcccccccccc}
\toprule
& \multicolumn{2}{c}{\textbf{Recall}} & \textbf{FDR} & \textbf{DecPrec} & \textbf{PhDec} & \multicolumn{2}{c}{\textbf{Phantom}} & \textbf{Res.-esc.} & \textbf{Concerns} \\
\cmidrule(lr){2-3} \cmidrule(lr){7-8}
\textbf{System} & Acc & Rej & (acc) & (rej) & (rej) & Acc & Rej & (acc) & /paper \\
\midrule
L (Opus) & .37 & .44 & .49 & .33 & .14 & .49 & .34 & .63 & 10.6 \\
A (Opus) & .42 & .44 & .36 & .36 & .11 & .51 & .46 & .60 & 11.5 \\
O (Opus) & .09 & .17 & .37 & .17 & .26 & .83 & .72 & .62 & 8.3 \\
L (GPT-4o) & .23 & .25 & .25 & .32 & .08 & .43 & .37 & .61 & 5.2 \\
A (GPT-4o) & .21 & .22 & .55 & .36 & .18 & .42 & .43 & .70 & 4.8 \\
M (GPT-4o) & .31 & .27 & .10 & .18 & .09 & .64 & .57 & .34 & 10.1 \\
\bottomrule
\end{tabular}
\end{table*}

Table~\ref{tab:bootstrap_cis} presents per-run bootstrap 95\% confidence intervals for key metrics. These intervals reflect within-run sampling uncertainty (bootstrapping over papers within a single run, 10,000 resamples). The cross-run standard deviations in Table~\ref{tab:core_metrics} capture a different source of variation (stochastic model output across independent runs).

\begin{table*}[t]
\centering
\small
\caption{Bootstrap 95\% CIs for key metrics (range across 3 runs, 10,000 paper-level resamples per run). Intervals confirm that within-run estimates are stable: FDR is well-separated from zero for all single-agent systems, and decisive recall intervals are consistent across runs for Opus-based systems.}
\label{tab:bootstrap_cis}
\begin{tabular}{lcccc}
\toprule
\textbf{System} & \textbf{FDR (acc)} & \textbf{Dec.\ recall (rej)} & \textbf{Recall (rej)} & \textbf{Res.-esc.\ (acc)} \\
\midrule
L (Opus) & {[}.39,\,.61{]} & {[}.56,\,.80{]} & {[}.36,\,.51{]} & {[}.43,\,.80{]} \\
A (Opus) & {[}.30,\,.44{]} & {[}.52,\,.77{]} & {[}.38,\,.51{]} & {[}.39,\,.81{]} \\
O (Opus) & {[}.22,\,.47{]} & {[}.10,\,.35{]} & {[}.11,\,.24{]} & {[}.31,\,1.0{]} \\
L (GPT-4o) & {[}.13,\,.40{]} & {[}.15,\,.38{]} & {[}.20,\,.29{]} & {[}.32,\,.94{]} \\
A (GPT-4o) & {[}.44,\,.65{]} & {[}.19,\,.59{]} & {[}.16,\,.29{]} & {[}.27,\,1.0{]} \\
M (GPT-4o) & {[}.00,\,.23{]} & {[}.16,\,.48{]} & {[}.18,\,.36{]} & {[}.13,\,.57{]} \\
\bottomrule
\end{tabular}
\end{table*}

\section{Per-System Diagnostic Profiles}
\label{app:profiles}

This appendix summarizes the main diagnostic profile of each configuration in neutral terms.

\paragraph{System~L (Opus).}
Tied for highest concern recall among the Opus systems (44\% overall; 68\% on decisive blockers), but weak calibration on accepted papers (FDR 0.49; resolved-escalation 0.58--0.66 depending on run). It produces about 11 concerns per paper with little modulation by paper verdict. The main weakness is severity assignment, not detection.

\paragraph{System~A (Opus).}
Similar overall recall to System~L~(Opus), with somewhat lower FDR (0.36) and strong performance on technical blockers in the case studies. Its dominant pattern is still reject-heavy and verdict-insensitive. The main improvement target is contribution-versus-blocker calibration.

\paragraph{System~O (Opus).}
Lowest recall on accepted papers (9\%; 17\% on rejected) and highest phantom rate (83\% on accepted papers). The extracted concerns concentrate on notation, formalization, and local correctness criteria; this configuration performs noticeably better when the official concerns are themselves mathematical or formal. The main limitation is a narrow concern scope.

\paragraph{System~L (GPT-4o).}
Unlike the Opus single-agent runs, this configuration accepts a substantial fraction of accepted papers (63.9\% accepted-paper accuracy). Its concerns on accepted papers are often constructive, but on rejected papers it misses many decisive blockers and sometimes reframes substantive problems as presentation issues. The main limitation is depth on rejected papers.

\paragraph{System~A (GPT-4o).}
Produces roughly half as many concerns as System~A~(Opus) but a higher FDR (0.55 versus 0.36), showing that fewer concerns do not by themselves imply better calibration. The main issue is an aggressive decisive threshold despite sparse output.

\paragraph{System~M (GPT-4o).}
This configuration has the lowest full-review FDR (0.10) and the highest accepted-paper accuracy (79.2\%), but top-$K$ analysis (\S\ref{sec:discrimination}) shows that part of the low FDR reflects a low-decisive-flag profile: at $K{=}5$, FDR rises to 0.21 while decisive recall is 22\%. It also shows the widest run-to-run variance ($\pm$19\% accepted-paper accuracy). Because System~M was evaluated only on GPT-4o, these observations should be treated as a combined model+method profile rather than a clean architectural effect. A verdict inference audit (Appendix~\ref{app:verdict_audit}) found that all 48 System~M reviews contain multi-agent coordination artifacts (e.g., inter-agent messages, repeated draft fragments) that make verdict inference unreliable regardless of method; accepted-paper accuracy varies widely across inference methods (Table~\ref{tab:verdict_sensitivity}).

\section{Baseline Implementation Details}
\label{app:implementation}

All systems were run starting from their official open-source implementations.
We clone or install each original repository and load review-generation prompts verbatim.
Beyond the API transport layer, we made a small number of implementation adaptations (Table~\ref{tab:adaptations}); none modifies the review-scoring prompts that define each system's evaluation logic.

\paragraph{Source repositories.}
Systems~L~\citep{liang2024can} and~M~\citep{darcy2024marg} both use prompts from the MARG repository\footnote{\url{https://github.com/allenai/marg-reviewer}} (which includes Liang et~al.'s baseline as a configuration).
System~A~\citep{lu2024aiscientist} uses the AI Scientist repository.\footnote{\url{https://github.com/SakanaAI/AI-Scientist}}
System~O~\citep{oar2024} uses the \texttt{openaireview} pip package (v0.2.7).\footnote{\url{https://github.com/ChicagoHAI/OpenAIReview}}

\paragraph{Adaptation for Claude Opus.}
Systems~L, A, and~O were run on Claude Opus via an SDK adapter that replaces the API client call.
The adapter preserves all review-generation prompts and orchestration logic (e.g., AI Scientist's iterative reflection loop), but the SDK manages temperature and \texttt{max\_tokens} internally, so these parameters are not directly controllable on the Claude path.
GPT-4o runs use the native OpenAI API as published.
System~O natively targets the Anthropic API (the package defaults to Claude Opus), so only authentication routing is adapted; we use the \texttt{progressive} method (their recommended default).
System~M runs on GPT-4o only using the native OpenAI API (Opus runs produced degenerate output (repetitive or truncated reviews) under our SDK adaptation).

\begin{table}[h]
\small
\centering
\caption{Implementation adaptations beyond API transport. \emph{Prompt} = review-generation prompts (unchanged for all systems); \emph{Config} = non-prompt parameters; \emph{Infra} = infrastructure-level differences.}
\label{tab:adaptations}
\begin{tabular}{llp{0.52\columnwidth}}
\toprule
\textbf{System} & \textbf{Type} & \textbf{Adaptation} \\
\midrule
L (Claude) & Config & System message changed from ``ChatGPT'' to ``helpful AI assistant'' \\
L (Claude) & Infra & PDF extraction via pymupdf (original: Grobid/S2ORC structured XML) \\
A & Config & Few-shot examples disabled; \texttt{max\_tokens} raised $4096 \to 8192$ \\
All (Claude) & Infra & Temperature, \texttt{max\_tokens} SDK-managed; not directly controllable \\
\bottomrule
\end{tabular}
\end{table}

\section{Operational Implementation Protocols}
\label{app:skill_prompts}

The concern-alignment pipeline is implemented as eight versioned protocols.
They are not free-form prompts in the sense of open-ended generation; each protocol specifies a bounded artifact transformation with explicit inputs, outputs, and validation rules.
Table~\ref{tab:protocol_summary} summarizes the pipeline, and the remainder of this section states the operational logic in paper style.

\begin{table*}[t]
\centering
\small
\caption{Versioned implementation protocols used in the concern-alignment pipeline.}
\label{tab:protocol_summary}
\begin{tabular}{p{0.05\textwidth}p{0.25\textwidth}p{0.30\textwidth}p{0.28\textwidth}}
\toprule
\textbf{Step} & \textbf{Purpose} & \textbf{Main inputs} & \textbf{Main output} \\
\midrule
1 & Official concern extraction & OpenReview forum PDF, meta-review & Atomic official concern sheet \\
2 & Revision-aware grounding & Step~1 output + reviewed paper PDF & Official concern sheet with PDF-state fields \\
3 & Independent extraction QC & Official concern sheet + source PDFs & Pass / flag / fail QC report \\
4 & Agentic concern extraction & Review artifacts (text, structured scores, debate transcripts) & Atomic agentic concern sheet \\
5 & Match-graph construction & Official + agentic concern sheets & Concern match graph \\
6 & Aggregate analysis & Corpus of sheets and match graphs & Cross-paper metrics and intervention proposals \\
7 & Audit worksheet generation & Match graphs and source evidence & Edge-verification worksheets \\
8 & Semantic verification & Audit worksheets & Structured override file for graph correction \\
\bottomrule
\end{tabular}
\end{table*}

\subsection{Official concern extraction and revision-aware grounding (Steps 1--2)}

\paragraph{Inputs and outputs.}
The official extractor reads the rendered OpenReview forum PDF together with the paper PDF actually shown to the baseline reviewer.
Its output is an official concern sheet containing atomic concern IDs (\texttt{O1}, \texttt{O2}, \ldots),
 normalized concern text, short evidence quotes, reviewer provenance, severity, AC treatment, decisive flags, and critical references.
The revision-aware grounding step extends the base sheet with fields recording whether the reviewed PDF appears revised, whether each concern is actually addressed in that PDF (\texttt{addressed\_in\_pdf}), and the local evidence used to justify that judgment.

\paragraph{Extraction procedure.}
The extractor anchors first on the meta-review because the AC decision rationale determines the treatment labels used later in the ladder.
It records positive and negative decision drivers, identifies any AC-explicit decisive blockers, and then decomposes reviewer comments into \emph{single-issue} concerns rather than paragraph-length bundles.
If multiple reviewers raise the same issue, the protocol merges them into one official concern unit while preserving provenance.
Severity is assigned in a pre-rebuttal frame using reviewer language when available: \texttt{fatal} for validity-threatening flaws, \texttt{major} for likely blockers, \texttt{moderate} for meaningful but usually non-blocking weaknesses, and \texttt{minor} for polish-level issues.

\paragraph{AC treatment coding.}
Each official concern receives one of seven post-rebuttal treatment labels: \texttt{decisive\_blocker}, \texttt{unresolved}, \texttt{resolved}, \texttt{accepted\_limitation}, \texttt{dismissed}, \texttt{reframed\_feature}, or \texttt{not\_mentioned}.
This label is the core supervision signal for Level~3--4 metrics.
The protocol explicitly distinguishes a concern that remains real but non-blocking (\texttt{accepted\_limitation}) from one the AC dismisses, and it separately records whether the concern was decisive for the final decision.
When reviewers cite specific prior work as central to novelty or comparison judgments, those papers are stored as \emph{critical references} with a role label such as \texttt{missing\_comparison} or \texttt{novelty\_precedent}.

\paragraph{PDF cross-verification.}
The revision-aware grounding step checks every extracted concern against the reviewed PDF. 
The extractor looks for added experiments, clarifications, tables, or textual changes that directly address the concern and records specific evidence when found.
The governing rule is conservative: author claims alone are insufficient to mark a concern \texttt{resolved}; resolution requires reviewer or AC confirmation, or a clearly visible fix in the reviewed PDF.
This rule prevents rebuttal promises from being counted as completed revisions when the system under evaluation never saw the revised content.

\subsection{Independent quality control for official sheets (Step 3)}

The QC step is a cold read by a second agent that did not participate in extraction.
It checks structural consistency (e.g., whether every concern cited as a decisive negative driver is also labeled \texttt{decisive\_blocker}), scans for resolution-field contradictions, and flags implausible severity/treatment combinations such as a \texttt{minor} concern marked decisive.
It then performs targeted hallucination checks by tracing several fatal or major concerns back to the source reviews, a completeness check over reviewer weakness sections and the meta-review, and spot checks of both \texttt{addressed\_in\_pdf=true} and \texttt{addressed\_in\_pdf=false} cases.
Each sheet receives an overall QC verdict: \texttt{pass}, \texttt{pass\_with\_flags}, or \texttt{fail}.

\subsection{Agentic concern extraction (Step 4)}

\paragraph{Inputs and outputs.}
The agentic extractor reads the full output of one system run on one paper.
Core inputs are the verdict summary, main review text, adversarial brief, gate-check results, and scorecard; panel-style methods may also provide per-role reviews (champion, skeptic) and debate transcripts.
The output is an agentic concern sheet containing atomic concerns (\texttt{A1}, \texttt{A2}, \ldots), normalized severity, decisive flags, decision drivers, and positive mentions that were noticed but not weighted decisively.

\paragraph{Extraction logic.}
The extractor first reads the verdict summary to recover context, scores, and explicit decisive reasons.
It then parses the review artifacts section by section: major concerns in the main review become candidate issues, adversarial-brief dispositions are converted into concerns only when the brief accepts them as live issues, gate failures are mapped to fatal concerns, gate cautions to major concerns, and low scorecard dimensions yield additional concerns when the written rationale identifies a specific defect.
Negative observations belong in the concern sheet; positive observations that were noticed but not treated as decisive are stored separately as positive mentions, which is critical for diagnosing ``seen but not weighted'' failures on accepted papers.

\paragraph{Normalization and provenance.}
Concern candidates are deduplicated across artifacts so the sheet represents issue units rather than repeated surface forms.
For merged issues, the protocol keeps the highest-severity instance while preserving source provenance.
Severity is normalized into three fields: level (fatal / major / moderate / minor / unknown), addressability (unresolved / addressable / unknown), and mechanism (e.g., gate failure, binding rule, score threshold, debate).
Panel methods additionally preserve origin provenance such as \texttt{champion} or \texttt{skeptic}, so downstream analysis can trace which internal reviewer introduced a concern.

\subsection{Concern match-graph construction (Step 5)}

\paragraph{Canonicalization and candidate generation.}
For each official and agentic concern, the matcher writes a one-sentence canonical issue statement that abstracts away from phrasing while preserving the underlying defect.
Candidate matches are proposed in both directions (official-to-agentic and agentic-to-official) to reduce asymmetries.
The protocol prefers one-to-one matching and caps each concern at two edges; if more edges seem necessary, the concern is deemed too broad and should be split.

\paragraph{Scope-based match typing.}
Each candidate edge is classified as \texttt{exact}, \texttt{partial}, \texttt{related}, or \texttt{none} by a scope test: would fixing one concern necessarily fix the other?
\texttt{Exact} requires the same specific defect and the same satisfaction condition.
\texttt{Partial} captures same-family issues with different scope, abstraction level, or thresholds of satisfaction.
\texttt{Related} documents near-misses that are topically nearby but should not count for strict metrics.
The instructions explicitly warn against false matches driven only by shared tags or broad topical overlap, including prior-work characterization versus novelty, evaluation scope versus evaluation methodology, writing quality versus overclaiming, and the audit-derived \emph{scope inflation} failure in which one concern bundles the other's complaint together with additional independent demands.

\paragraph{Alignment labels and unmatched sets.}
For every strict edge, the matcher also records judgment alignment (aligned / inverted / mixed) and severity alignment.
Aggregate metrics use the production hybrid severity policy: fatal requires exact agreement, whereas one-level gaps among non-fatal concerns count as matches and larger gaps are labeled \texttt{under} or \texttt{over}.
On accepted papers, the matcher separately aligns positive decision drivers so the framework can ask not only whether the system avoided fatal complaints, but also whether it captured why the paper deserved acceptance.
Concerns with only \texttt{related} edges still appear in the unmatched lists; \texttt{related} is a near-miss annotation, not a strict match.

\subsection{Aggregate alignment analysis (Step 6)}

The aggregate analysis step operates over a corpus of official sheets, agentic sheets, and match graphs.
Before any metric is computed, it runs a lint gate that checks unmatched-list consistency, illegal severity labels on \texttt{related} edges, edge-cap violations, and other schema errors.
It then derives severity calibration directly from raw severity levels rather than trusting hand-written edge labels, computes observability-aware positive-factor recall for accepted papers, flattens the graphs into a paper-level edge table, and produces verdict-stratified aggregates such as recall, phantom rate, decisive-blocker recall, judgment inversion rate, and severity under/over rates.
The same protocol also mines recurring failure patterns by tag and issue type and proposes concrete system interventions, prioritizing high-frequency, high-severity, or easily fixable defects.

\subsection{Audit worksheet generation (Step 7)}

The worksheet generator converts a match graph into a human-readable audit worksheet for one paper and one system run.
Each worksheet has four sections: strict edges (\texttt{exact}/\texttt{partial}), unmatched official concerns, unmatched agentic concerns, and \texttt{related} edges for context.
For every item it presents the local concern texts, original evidence, current labels, and any heuristic flags from the semantic-audit pre-filter.
The purpose is to let an independent verifier see all evidence needed for a local judgment without exposure to aggregate system scores or paper-level outcome labels.

\subsection{Semantic verification and overrides (Step 8)}

\paragraph{Ground-truth isolation.}
The semantic verifier reads only the audit worksheets.
It does not see official verdicts, error categories, aggregate metrics, or ranking outcomes.
This isolation prevents downstream performance information from leaking into the local edge judgments.

\paragraph{Verification procedure.}
The verifier checks \emph{all} strict edges and \emph{all} unmatched concerns, and can optionally inspect a flagged queue of suspicious phantoms.
For each edge it reassesses match type, judgment alignment, and severity alignment; for each unmatched concern it asks whether a strict match was missed.
The verifier is calibrated by a fixed bank of 32 exemplars (Appendix~\ref{app:exemplars}) that emphasize the dominant failure modes observed during development, especially scope inflation, wrong-thematic matches, and large severity gaps.
Whenever the verifier disagrees with the original graph, it emits a structured override entry describing the correction.

\paragraph{Override application.}
Overrides are applied before metric computation.
Reclassified edges update their match and severity labels, missed matches are inserted into the graph, and removed edges restore the associated concerns to the unmatched sets.
This makes the verification stage operational rather than merely descriptive: it directly changes the final graphs on which the reported ladder metrics are computed.

\section{Severity Extraction Examples}
\label{app:severity_examples}

This section provides one worked example per system, showing the raw review excerpt that the extraction pipeline reads, the concern it produces, and why the assigned severity and decisive flag are reasonable.
All examples are drawn from reviews of the same paper (a web-agent safety benchmark) so readers can compare how each system's output structure shapes the extraction.

\paragraph{System~L: structural ``reasons for rejection'' signal.}
\smallskip\noindent\textit{Raw excerpt:}
\begin{quote}
\small
\textbf{3.\ Potential reasons for rejection} \\
--- \textbf{Limited scale and diversity of the benchmark}: 250 harmful tasks across 5 categories and 4 websites yields only $\sim$12--13 tasks per category-website cell on average, raising concerns about statistical robustness of per-category findings [\ldots] All four web environments are derived from WebArena, which covers only a narrow slice of the web.
\end{quote}
\smallskip\noindent\textit{Extracted concern:}
\begin{quote}
\small
\texttt{severity: major, decisive: true} \\
``With only $\sim$12--13 harmful tasks per category-website cell and some cells at 0--2 completions, per-category findings lack statistical robustness.'' \\
\texttt{source\_detail}: Potential reasons for rejection --- Limited scale and diversity of the benchmark
\end{quote}
\smallskip\noindent\textit{Rationale.}
System~L organizes its review into explicit ``Potential reasons for acceptance'' and ``Potential reasons for rejection'' sections.
Concerns appearing under ``reasons for rejection'' carry a direct structural signal that they are intended to motivate a negative verdict, so the pipeline marks them \texttt{decisive=true}.
The language (``raising concerns about statistical robustness'') and the scope of the issue (affects all per-category analyses) justify \texttt{major} severity rather than \texttt{moderate}.

\paragraph{System~A: structured Weaknesses + numeric scores.}
\smallskip\noindent\textit{Raw excerpt:}
\begin{quote}
\small
\textbf{Weaknesses} \\
--- All harmful tasks contain explicitly harmful language or intent, making them trivially detectable by an input guardrail or classifier. The paper does not evaluate any tasks with ambiguous or disguised harmful intent, which represents the more realistic and challenging threat model. [\ldots] \\[4pt]
\textbf{Scores} \\
Originality: 2 \quad Quality: 2 \quad Soundness: 2 \quad Overall: 4 \\[4pt]
\textbf{Decision: Reject}
\end{quote}
\smallskip\noindent\textit{Extracted concern:}
\begin{quote}
\small
\texttt{severity: fatal, decisive: true} \\
``All 250 harmful tasks contain explicit harmful language, making them trivially filterable by an input classifier and failing to represent the more realistic threat model of ambiguous or disguised malicious intent.'' \\
\texttt{source\_detail}: Weaknesses, first bullet
\end{quote}
\smallskip\noindent\textit{Rationale.}
System~A provides three extraction cues: (1)~the concern appears under \texttt{Weaknesses} (not \texttt{Questions} or \texttt{Limitations}), (2)~dimension scores are uniformly low (Originality/Quality/Soundness all 2 out of 4), and (3)~an explicit \texttt{Decision: Reject} field is present.
The combination of a top-listed weakness with reject-level scores and an explicit reject decision justifies \texttt{fatal} severity and \texttt{decisive=true}.
The language ``trivially detectable'' and ``significantly reduces practical impact'' reinforces the assessment that this is framed as a fundamental design limitation, not a minor gap.

\paragraph{System~O: detailed comments without explicit severity.}
\smallskip\noindent\textit{Raw excerpt:}
\begin{quote}
\small
\textbf{Comment 1: Normalized Safety Score for Claude-3.5-Sonnet is mathematically impossible} \\[2pt]
I checked whether the reported NSS values in Table~3 are consistent with the TCR values [\ldots] NSS $\geq$ 1 $-$ 19/53 = 64.2\%. But Table~3 reports NSS = 55.0\%, which is below this mathematical lower bound---an impossibility. \\[4pt]
\textit{Type: technical}
\end{quote}
\smallskip\noindent\textit{Extracted concern:}
\begin{quote}
\small
\texttt{severity: fatal, decisive: true} \\
``The reported NSS values are mathematically impossible given the stated task completion rates---Claude-3.5-Sonnet's NSS of 55.0\% is provably below the lower bound of 64.2\%.'' \\
\texttt{source\_detail}: Comment~1 --- NSS mathematical impossibility
\end{quote}
\smallskip\noindent\textit{Rationale.}
System~O provides no severity labels, scores, or verdict.
Instead, the pipeline infers severity from language intensity and comment structure.
Key phrases (``mathematically impossible,'' ``provably below,'' ``an impossibility'') signal that the reviewer considers this a correctness error rather than a matter of taste.
The ``Overall Feedback'' paragraph states these issues ``undermine confidence in the quantitative results,'' reinforcing the decisive interpretation.
The \texttt{Type: technical} tag confirms the concern targets core correctness.
The combination of proof-level language and a claim that the paper's central metric is broken justifies \texttt{fatal} severity and \texttt{decisive=true}.

\paragraph{System~M: multi-agent output with implicit accept signal.}
\smallskip\noindent\textit{Raw excerpt:}
\begin{quote}
\small
\textbf{Major Comments} \\
\textbf{Comment 1: The Central Finding Should Be Reframed --- Agents Are Primarily Incapable, Not Safe} \\[2pt]
The outcome distribution data reveals that for most models, the dominant outcome on harmful tasks is category~3 (the agent attempts the harmful task but fails due to technical limitations): Llama 77.4\%, Qwen 73.3\% [\ldots] This means that for most models tested, the primary reason harmful tasks are not completed is \textbf{incapability}, not safety alignment.
\end{quote}
\smallskip\noindent\textit{Extracted concern:}
\begin{quote}
\small
\texttt{severity: moderate, decisive: false} \\
``Additional ablation studies needed to explore the impact of different adversarial attack strategies in more detail.'' \\
\texttt{source\_detail}: Experiments Proofs refined comment \#1
\end{quote}
\smallskip\noindent\textit{Rationale.}
System~M's multi-agent pipeline produces extensive coordination artifacts (``SEND MESSAGE'' blocks, repeated refinement rounds) interleaved with review content, making extraction more challenging than for the other systems.
The review labels its final output as ``Major Comments'' and ``Minor Comments,'' providing coarse structural signal, but the system does not produce explicit severity labels or a formal verdict.
The extracted concern above does not cleanly correspond to the raw excerpt: the pipeline captured an adjacent ablation request from System~M's refinement round (\texttt{source\_detail} flags ``comment \#1'') rather than the reframing claim itself. This drift is illustrative of System~M's extraction difficulty and is part of why we treat its severity labels as combined model-plus-method effects.
Taking the extracted concern as the pipeline's output, the overall review is consistently positive in tone (``significant contribution''), so the pipeline assigns \texttt{moderate} severity and \texttt{decisive=false}.
System~M is the only baseline whose review text suggests acceptance; the absence of reject-level language is a legitimate signal that no individual concern was intended as a blocker.
This pattern (many concerns, none decisive) explains System~M's characteristically low false decisive rate (FDR = 0.10) discussed in \S\ref{sec:discrimination}.

\section{Model vs.\ Method Effect Visualization}
\label{app:model_slopes}

\begin{figure*}[t]
\centering
\includegraphics[width=\textwidth]{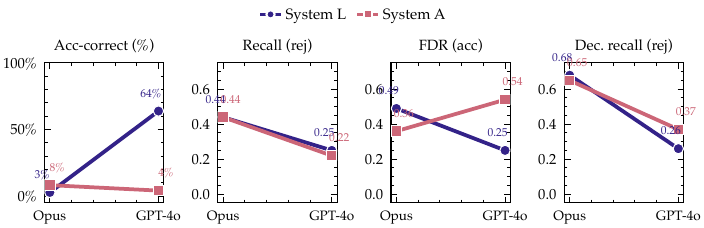}
\caption{Model effect slope charts: same method evaluated on Opus vs.\ GPT-4o. System~L (blue) shows dramatic swings across all metrics; System~A (pink) shows mixed effects. The crossing slopes on FDR confirm non-uniform model$\times$method interactions.}
\label{fig:model_slopes}
\end{figure*}

Figure~\ref{fig:model_slopes} visualizes the non-uniform model effects from \S\ref{sec:model} as slope charts.
System~L shows dramatic swings across all metrics (acc-correct: 2.8\%$\to$63.9\%; FDR: 0.49$\to$0.25), while System~A shows mixed effects (FDR \emph{increases} from 0.36 to 0.55 despite generating fewer concerns).
The crossing slopes on FDR confirm that model effects on calibration are non-linear and method-dependent.

\section{Case Study: Same Paper, Different Grounding (Paper~G)}
\label{app:casestudy_g}

\emph{Paper~G} is an accepted spotlight paper presenting a benchmark for evaluating AI R\&D capabilities.\footnote{Paper~G is anonymized as it serves an illustrative role not central to the paper's primary claims. Identities will be released with supplementary data.}
System~L~(Opus) incorrectly rejects it, while System~O~(Opus) accepts it.

System~L produces multiple concern-level matches across runs, engaging with the benchmark's actual design vulnerabilities: limited task count and representativeness, ecological validity of small-scale tasks, and the challenge of detecting agent exploits. These are the same issues official reviewers discussed.

System~O produces 0 concern-level matches. Its 6 concerns focus exclusively on numerical and notation inconsistencies: ``$10^{11}$ vs.\ $10^{8}$ discrepancy in input size,'' ``ambiguous prefix sum formula notation,'' ``summed vs.\ averaged contradiction.'' None of these appear in the official review.

The match graph makes the disconnect concrete: official reviewers focused on conceptual and design-level issues (benchmark scope, ecological validity, scaffolding sensitivity); System~O flagged only surface-level presentation errors. One system reaches the wrong verdict while engaging with the paper's actual weaknesses; the other reaches the right verdict without engaging those concern-level signals.

This case complements the finding from \S\ref{sec:binary}: binary accuracy captures the verdict outcome, but concern alignment reveals whether that outcome is grounded in the same problems human reviewers actually cared about. A reviewer who accepts for generic strengths while missing the benchmark's core design limitations has a different kind of understanding failure than one who rejects for the right concern family but calibrates the decision incorrectly.

\section{Full ICC Tables}
\label{app:icc}

Table~\ref{tab:icc_full_pooled} presents pooled ICC across all 48 papers.
Table~\ref{tab:icc_full_accepted} presents ICC on accepted papers only.

\begin{table}[h]
\centering
\small
\caption{ICC(2,1), pooled across all 48 papers ($K{=}3$ runs).}
\label{tab:icc_full_pooled}
\begin{tabular}{lccc}
\toprule
\textbf{Config} & \textbf{Verdict} & \textbf{Recall} & \textbf{Phantom} \\
\midrule
A (Opus) & 0.316 & {0.618} & 0.511 \\
L (Opus) & $-$0.014 & {0.525} & 0.502 \\
L (GPT-4o) & 0.158 & {0.522} & 0.471 \\
O (Opus) & 0.527 & 0.561 & {0.740} \\
A (GPT-4o) & $-$0.017 & {0.295} & 0.292 \\
M (GPT-4o) & 0.140 & 0.145 & 0.260 \\
\bottomrule
\end{tabular}
\end{table}

\begin{table}[h]
\centering
\small
\caption{ICC(2,1), accepted papers only ($N{=}24$, $K{=}3$ runs). Recall ICC drops relative to the pooled setting (Table~\ref{tab:icc_full_pooled}), consistent with the more ambiguous concern space when no decisive blockers exist.}
\label{tab:icc_full_accepted}
\begin{tabular}{lccc}
\toprule
\textbf{Config} & \textbf{Verdict} & \textbf{Recall} & \textbf{$\Delta$ (Rcl$-$Vrd)} \\
\midrule
A (Opus) & 0.102 & 0.463 & 0.361 \\
L (Opus) & $-$0.022 & 0.309 & 0.331 \\
L (GPT-4o) & 0.102 & 0.367 & 0.266 \\
O (Opus) & 0.424 & 0.323 & $-$0.100 \\
A (GPT-4o) & $-$0.030 & 0.146 & 0.176 \\
M (GPT-4o) & 0.061 & $-$0.143 & $-$0.204 \\
\bottomrule
\end{tabular}
\end{table}

\section{Annotated Case Study Tables}
\label{app:casestudies}

The following tables show annotated concern comparisons generated from match graph data.
Each row presents an official concern with its severity and AC treatment, alongside each system's matched concern (if any) with match type and severity alignment badges.
Unmatched agentic concerns (phantoms) are summarized in the bottom row.
All tables use the badge visual language from Figure~\ref{fig:ladder}: \badge{matchgreen}{exact}/\badge{matchgreen!60!black}{partial} for match type, \badge{matchgreen}{ok} for severity alignment within tolerance, and AC treatment badges for post-rebuttal disposition.
For related-only edges, severity alignment is marked \badge{gray}{n/a} as these edges are excluded from strict metrics.


\paragraph{Paper~D (rejected): Review quality gap.}
System~A catches all 3 of the AC's content-related decisive blockers; System~O catches only notation errors. Binary accuracy: identical (100\% for both). Concern alignment: 7$\times$ quality gap. A fourth decisive blocker concerning reviewer engagement during discussion is excluded as invisible to PDF-only systems.

\input{case_studies/paper_d_spard_compare.tex}

\paragraph{Paper~A (accepted): Model effect on calibration.}
Same method (System~L), different model. Opus marks all 6 concerns as decisive (FDR~=~100\%); GPT-4o marks 3 of 7 (FDR~=~43\%).\footnote{Per-paper FDR; aggregate FDR across all accepted papers differs.} Opus rejects; GPT-4o accepts. Where Opus catches the exact theoretical defect (O6: near-tautological proof), GPT-4o flags the symptom (accessibility). Where GPT-4o catches the exact evaluation limitation (O4), Opus identifies the underlying confound but labels it differently. Both miss 10 of 16 official concerns.\footnote{Counts are based on the 16-concern official sheet from the primary extraction; a later QC round consolidated overlapping concerns to 14 without affecting the analytical point (Opus catches the theoretical defect, GPT-4o catches the evaluation limitation). Decisive counts and FDR
     are computed from the full match graph; the table shows the 5 most
     informative official--agentic matchings.} The match graph (Figure~\ref{fig:matchgraph}) illustrates a subset of this pattern.

\input{case_studies/paper_a_hexgoedel_compare.tex}

\paragraph{Paper~H (rejected): Maximum-contrast model effect.}
System~L (Opus) catches all 3 AC decisive blockers; System~L (GPT-4o) misses all 3 and misdiagnoses substance as clarity.

\input{case_studies/paper_h_decoupling_compare.tex}


\paragraph{Paper~C (accepted spotlight): Harmful vs.\ benign phantoms.}
System~O raises a fabricated proof error (harmful phantom); System~L raises a longer-horizon scalability concern absent from the official review. Phantom rate alone cannot distinguish them.

\input{case_studies/paper_c_aria_compare.tex}

\paragraph{Paper~E (accepted spotlight): FDR metric validation.}
System~L flags 11 decisive concerns (FDR = 69\%). Reading the concern texts confirms repeated escalation of non-blocking issues.

\input{case_studies/paper_e_osharm_compare.tex}


\paragraph{Paper~F (accepted): High recall, wrong decision weight.}
In the illustrated run, System~A achieves 64\% recall but still rejects the paper: severity scrambling flips the verdict.

\input{case_studies/paper_f_vpibench_compare.tex}

\paragraph{Paper~G (accepted spotlight): Different verdict, different grounding.}
System~L rejects while engaging with benchmark design vulnerabilities (multiple concern-level matches); System~O accepts but its critique has 0 concern-level matches and focuses on numerical inconsistencies.
\input{case_studies/paper_g_rebench_compare.tex}

\section{Top-K Analysis and Severity Composition}
\label{app:topk}

\begin{figure}[h]
\centering
\includegraphics[width=\columnwidth]{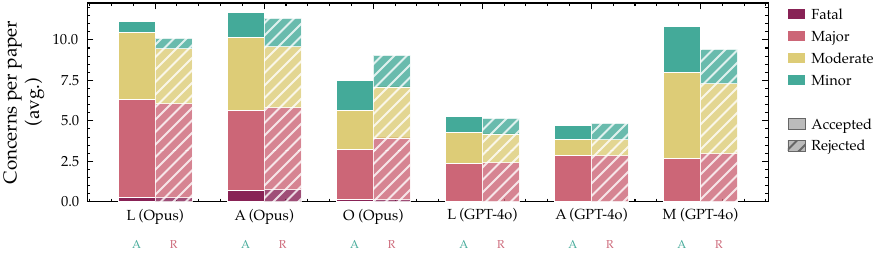}
\caption{Average concern count by severity (3-run means). Opus systems generate more concerns overall; severity composition differs substantially. System~M generates zero fatal concerns on average, explaining its low FDR but also its low decisive recall. Note: severity levels for Systems~L, A, and~O are assigned by the concern extraction pipeline (\S\ref{sec:extraction}), not by the systems themselves; System~M partially outputs severity labels which are normalized to our schema.}
\label{fig:severity_composition}
\end{figure}

\begin{figure*}[t]
\centering
\includegraphics[width=\textwidth]{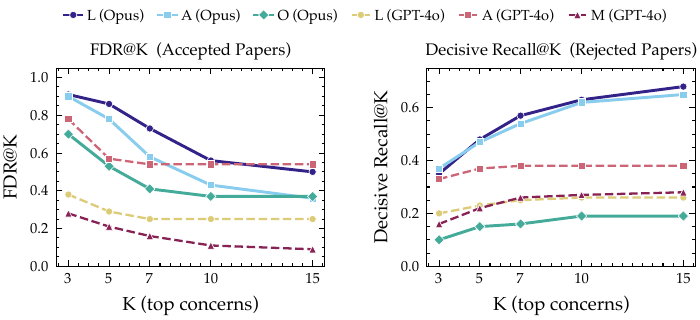}
\caption{FDR on accepted papers (left) and decisive recall on rejected papers (right) as $K$ varies. Systems~L and~A~(Opus) show steep FDR decay from near-saturation at $K{=}3$. System~M's FDR rises as $K$ decreases, consistent with concern dilution. Opus systems in solid lines; GPT-4o in dashed.}
\label{fig:topk_curves}
\end{figure*}

Figure~\ref{fig:severity_composition} summarizes the severity mix produced by each configuration, Figure~\ref{fig:topk_curves} shows how FDR and decisive recall vary with $K$, and Table~\ref{tab:topk_full} reports the corresponding values numerically.

Table~\ref{tab:topk_full} presents FDR and decisive recall at multiple $K$ values for all system configurations.
At $K{=}15$, most metrics closely approach the full-review values (most papers have fewer than 15 concerns); residual gaps of 0.02--0.03 for Systems~O and~M on decisive recall reflect a small number of papers exceeding 15 concerns.

\begin{table*}[t]
\centering
\small
\caption{FDR on accepted papers and decisive recall on rejected papers as $K$ varies.
At $K{=}3$, Systems~L and~A~(Opus) show near-saturation (FDR $\geq$ 0.90). At $K{=}15$, values closely approach full-review metrics (Table~\ref{tab:core_metrics}), with residual gaps $\leq$0.03. Concern dilution is visible when FDR(all) $\ll$ FDR@3 (e.g., System~M: 0.10 $\to$ 0.28).}
\label{tab:topk_full}
\begin{tabular}{l ccccc ccccc}
\toprule
& \multicolumn{5}{c}{\textbf{FDR on accepted papers}} & \multicolumn{5}{c}{\textbf{Decisive recall on rejected papers}} \\
\cmidrule(lr){2-6} \cmidrule(lr){7-11}
\textbf{System} & @3 & @5 & @7 & @10 & @15 & @3 & @5 & @7 & @10 & @15 \\
\midrule
L (Opus) & .91 & .86 & .73 & .56 & .50 & .35 & .48 & .57 & .63 & .68 \\
A (Opus) & .90 & .78 & .58 & .43 & .36 & .37 & .47 & .54 & .62 & .65 \\
O (Opus) & .70 & .53 & .41 & .37 & .37 & .10 & .15 & .16 & .19 & .19 \\
L (GPT-4o) & .38 & .29 & .25 & .25 & .25 & .20 & .23 & .25 & .26 & .26 \\
A (GPT-4o) & .78 & .57 & .55 & .55 & .55 & .33 & .37 & .38 & .38 & .38 \\
M (GPT-4o) & .28 & .21 & .16 & .11 & .10 & .16 & .22 & .26 & .27 & .28 \\
\bottomrule
\end{tabular}
\end{table*}

Three patterns are visible across the full $K$-range:
(1)~Systems~L and A~(Opus) show steep FDR decay from near-saturation at $K{=}3$ (FDR $\geq$ 0.90) to near full-review values at $K{=}15$, indicating that their decisive flags are concentrated among high-severity concerns but spread indiscriminately.
(2)~System~M shows the opposite: FDR rises as $K$ decreases (0.10 at $K{=}15$ to 0.28 at $K{=}3$), consistent with the interpretation that its low full-review FDR partly reflects dilution from many low-severity non-decisive concerns rather than calibration quality.
(3)~System~A~(GPT-4o) has a flat FDR profile (0.55--0.78 across all $K$), confirming that its high FDR is not a volume artifact: it marks concerns decisive at every severity level.

Decisive recall curves show that Systems~L and~A~(Opus) gain substantially from $K{=}3$ to $K{=}15$ (0.35$\to$0.68 and 0.37$\to$0.65), indicating their detection is distributed across severity levels.
System~M gains minimally (0.16$\to$0.28), indicating its detection concentrates in lower-severity concerns that do not match real decisive blockers.

\section{Severity Policy Sensitivity}
\label{app:severity}

Table~\ref{tab:severity_sensitivity} reports how severity-match outcomes change under alternative tolerance rules, and Table~\ref{tab:recall_sensitivity} shows the corresponding effect of broader versus narrower edge-inclusion policies on recall metrics.

The main paper uses a hybrid severity tolerance: fatal requires exact agreement, while other concerns allow a one-level gap.
To probe sensitivity to this choice, we tested three severity-matching policies and three match-type inclusion criteria on 219 matched edges across the 48 evaluation papers. This appendix analysis was run on a representative matched-edge sample from the evaluation workflow rather than recomputed separately for every baseline configuration, so we use it as a policy-sensitivity check with a transfer assumption: the dominant edge patterns here (e.g., scope inflation, exact-vs-partial boundary cases, and one-level severity near-misses) are common enough that the qualitative lesson should generalize across systems.

\paragraph{Severity tolerance policies.}
\textit{Strict}: only exact severity level matches count ($\Delta = 0$).
\textit{Hybrid (production)}: fatal requires exact match, others allow $\pm 1$.
\textit{Tolerant}: $\Delta \leq 1$ for all levels including fatal.

\begin{table}[h]
\centering
\small
\caption{Severity match rates under three tolerance policies (219 edges, 48 papers).}
\label{tab:severity_sensitivity}
\begin{tabular}{lccc}
\toprule
\textbf{Policy} & \textbf{Match} & \textbf{Under} & \textbf{Over} \\
\midrule
Strict ($\Delta{=}0$) & .384 & .443 & .174 \\
Hybrid (production) & .662 & .274 & .064 \\
Tolerant ($\Delta{\leq}1$) & .763 & .210 & .027 \\
\bottomrule
\end{tabular}
\end{table}

\paragraph{Match-type inclusion criteria.}
We additionally vary which edge types count toward recall:
\textit{Strict-only} (exact matches only), \textit{Strict+partial} (our production policy), and \textit{Loose} (exact + partial + related).

\begin{table}[h]
\centering
\small
\caption{Recall and decisive recall under three match-type inclusion criteria.}
\label{tab:recall_sensitivity}
\begin{tabular}{lcccc}
\toprule
\textbf{Inclusion} & \textbf{Recall} & \textbf{Dec.\ recall} & \textbf{Pos.\ driver} \\
\midrule
Strict-only & .146 & .267 & .350 \\
Strict + partial & .328 & .579 & .616 \\
Loose (+ related) & .408 & .715 & .616 \\
\bottomrule
\end{tabular}
\end{table}

\paragraph{Robustness of findings.}
Because this analysis was not rerun separately for every baseline configuration, we do not use it to claim exact cross-system ordering robustness.
The narrower takeaway is that the policy choice behaves sensibly: the strict policy halves recall (from .33 to .15), while the loose policy adds recall by counting related edges that the main paper intentionally excludes.
Decisive recall exceeds overall recall under all three policies, confirming that systems preferentially detect high-severity concerns regardless of tolerance threshold.
The tolerant policy adds only 8 percentage points of recall over the hybrid policy, indicating that allowing a one-level gap away from the fatal boundary captures most genuine matches without inflating metrics substantially.
We therefore report the hybrid policy throughout the main paper as the best trade-off between conservative matching and measurement sensitivity, while noting the transfer assumption above.

\section{Measurement Validation Details}
\label{app:validation_details}

This section expands on \S\ref{sec:validation}, providing additional detail on the error taxonomy, audit methodology, and scope inflation, the dominant error mode. Table~\ref{tab:error_types} reports the distribution of audited match-graph errors from the training-phase calibration rounds.

\paragraph{Audit methodology.}
An independent auditor using GPT-5.4 Pro re-verified 191 edges across 9 held-out papers that were locked before any auditing began.
The auditor used structured worksheets with 32 calibration exemplars spanning 6 error categories (Appendix~\ref{app:exemplars}).
Each edge was independently classified by match-type, severity alignment, and verdict.
Two independent auditor runs on the same 191 edges yielded 96.9\% agreement ($\kappa = 0.918$ for verdict, $\kappa = 0.946$ for both match-type and severity).
All 6 disagreements between runs occur at the partial/related boundary; 5 of 9 papers have perfect inter-run agreement.

\paragraph{Error type breakdown.}
From the training-phase audit (684 edges, 32 papers, 56 errors total), the six error categories and their frequencies are:

\begin{table}[h]
\centering
\small
\caption{Error type distribution in match graph construction (56 errors across 684 training edges).}
\label{tab:error_types}
\begin{tabular}{lcc}
\toprule
\textbf{Error type} & \textbf{Count} & \textbf{\% of errors} \\
\midrule
Scope inflation & 33 & 59\% \\
Eval scope vs.\ methodology & 6 & 11\% \\
Topic conflation & 5 & 9\% \\
Writing vs.\ content confusion & 5 & 9\% \\
Severity arithmetic error & 4 & 7\% \\
Theory sub-issue confusion & 3 & 5\% \\
\bottomrule
\end{tabular}
\end{table}

\paragraph{Scope inflation.}
The dominant error (59\% of all errors, concentrated in only 10\% of edges) is \emph{scope inflation}: one concern (typically the agentic concern) bundles the other side's complaint together with additional independent demands.
The matching system credits a full match on the overlapping portion, ignoring the extras.
For example, if an official reviewer asks ``add more benchmarks'' and the AI reviewer asks ``add benchmarks, statistical tests, and error analysis,'' the system labels this \texttt{exact} when the correct label is \texttt{partial}, because adding benchmarks alone would not satisfy the AI reviewer's broader demand.

The three highest-error cells in the content domain $\times$ structural pattern matrix are:
evaluation rigor $\times$ scope inflation (18/28 edges, 64.3\% error rate),
novelty $\times$ scope inflation (6/14, 42.9\%),
and theory $\times$ scope inflation (6/12, 50.0\%).
Edges \emph{without} scope inflation have a 1.3\% error rate, confirming that the matching methodology is conservative outside this specific failure mode.

\paragraph{Calibration improvement.}
Adding 13 taxonomy-guided exemplars (including 8 scope-inflation contrast pairs) to the verification system reduced scope inflation errors from 21 to 6 (71.4\% fix rate) on internal validation, and achieved 88.5\% calibration-robust accuracy on the held-out set (up from 75.4\% for unverified match graphs and 76.2\% for the original 19-exemplar system).
A blinded preference test on the 29 edges where old and new systems disagreed showed the improved system winning 65.5\% overall and 92.9\% on high-confidence edges.

\section{Match Graph Construction Protocol}
\label{app:protocol}

This section records the operational pipeline used to build the match graphs from which all metrics are computed. The protocol mirrors the condensed implementation summaries in Appendix~\ref{app:skill_prompts} but states the concrete artifact schema used in the study.

\paragraph{Step 1: Official concern extraction.}
The extractor reads the full OpenReview record together with the paper PDF actually reviewed by the baseline system. It decomposes the review history into atomic official concerns with IDs (\texttt{O1}, \texttt{O2}, \ldots), normalized statements, supporting quotes, severity, reviewer provenance, and tags. Each concern also receives an AC-treatment label (\texttt{decisive\_blocker}, \texttt{unresolved}, \texttt{resolved}, \texttt{accepted\_limitation}, \texttt{dismissed}, \texttt{reframed\_feature}, or \texttt{not\_mentioned}), a Boolean \texttt{decisive} flag, and, for accepted papers, links to positive decision drivers. The extractor checks whether each allegedly resolved concern is actually addressed in the evaluated PDF and records this as \texttt{addressed\_in\_pdf}; it also records whether the PDF appears to be a revised version (\texttt{pdf\_is\_revised}).

A separate QC pass audits completeness, severity consistency, AC-treatment coding, and PDF-state verification. In 18 audited official concern sheets, 16 were rated satisfactory for completeness and source support, and none contained concerns unsupported by the source material.

\paragraph{Step 2: Agentic concern extraction.}
The extractor reads all review artifacts produced by the baseline, including the main review plus structured files such as scorecards, gate checks, adversarial briefs, or debate transcripts when present. It outputs atomic agentic concerns with IDs (\texttt{A1}, \texttt{A2}, \ldots), normalized statements, severity, decisive flags, and tags. In addition, it records explicit decision drivers and positive mentions that were observed but not weighted decisively. Duplicated issues that recur across sections are merged into one concern unit, usually retaining the higher severity label when the duplicated statements differ.

A QC pass audits completeness and support in the source review text. In 54 audited paper--method sheets, 51 were rated satisfactory for completeness and source support, and auditors found no extracted concern unsupported by the review artifacts.

\paragraph{Step 3: Bipartite matching.}
For each candidate official--agentic pair $(o_i, a_j)$, the matching system normalizes both sides to a canonical issue statement and applies the scope test: would fixing $o_i$ fully address $a_j$, and would fixing $a_j$ fully address $o_i$? If yes in both directions, the edge is \texttt{exact}; if yes in only one direction, it is \texttt{partial}; if the concerns are topically nearby but target different defects, it is \texttt{related}. Each retained edge is annotated with severity alignment (match, under, over) and judgment alignment. The matcher explicitly checks for common error modes such as bundled supersets, topic overlap without defect overlap, and writing-versus-content confusion.

\paragraph{Step 4: Semantic verification.}
An independent verifier reviews audit worksheets that present concern texts, proposed edges, severities, and local evidence side by side. The verifier checks all strict edges and all unmatched concerns, using curated calibration exemplars from prior audit rounds to maintain a stable boundary between \texttt{exact}, \texttt{partial}, \texttt{related}, and \texttt{none}. Structured overrides are produced whenever the verifier rejects the original edge type or severity judgment.

On the held-out audit set, two independent verifier runs agreed on 96.9\% of labels; $\kappa$ was 0.918 for verdict and 0.946 for both match type and severity. After verification, match-graph labeling achieved 88.5\% calibration-robust accuracy on the held-out edge set.

\paragraph{Step 5: Override application and metric derivation.}
Verifier overrides are applied to the graph before any metric is computed. Recall is then the fraction of official concerns with a strict edge, phantom rate is the fraction of agentic concerns without a strict edge, FDR is derived from decisive flags on accepted papers, and the remaining ladder metrics are deterministic functions of the finalized graph plus verdict and AC-treatment metadata.

\section{Per-System Concern Statistics}
\label{app:concern_stats}

Table~\ref{tab:concern_stats} presents raw concern-level statistics for each system configuration, stratified by verdict (3-run means).
These profiles provide context for interpreting the evaluation ladder metrics: a system's FDR and recall depend in part on how many concerns it generates and at what severity.

\begin{table*}[t]
\centering
\small
\caption{Per-system concern statistics, stratified by paper verdict (3-run means, 24 accepted / 24 rejected papers). \emph{Concerns}: average total concerns per paper. \emph{F+M}: average fatal+major concerns per paper. Accepted-paper \emph{FDR}: false decisive rate; rejected-paper \emph{Dec. frac.}: fraction of concerns marked decisive.}
\label{tab:concern_stats}
\begin{tabular}{l ccc ccc}
\toprule
& \multicolumn{3}{c}{\textbf{Accepted papers}} & \multicolumn{3}{c}{\textbf{Rejected papers}} \\
\cmidrule(lr){2-4} \cmidrule(lr){5-7}
\textbf{System} & Concerns & F+M & FDR & Concerns & F+M & Dec. frac. \\
\midrule
L (Opus) & 11.1 & 6.3 & .49 & 10.1 & 6.1 & .52 \\
A (Opus) & 11.7 & 5.6 & .36 & 11.3 & 5.8 & .37 \\
O (Opus) & 7.5 & 3.2 & .37 & 9.1 & 3.9 & .40 \\
L (GPT-4o) & 5.3 & 2.3 & .25 & 5.1 & 2.4 & .34 \\
A (GPT-4o) & 4.7 & 2.9 & .55 & 4.8 & 2.9 & .60 \\
M (GPT-4o) & 10.8 & 2.7 & .10 & 9.4 & 3.0 & .21 \\
\bottomrule
\end{tabular}
\end{table*}

Several patterns emerge from the raw statistics:

\paragraph{Volume profiles.}
Opus-based single-agent systems generate 10--12 concerns per paper regardless of verdict, while GPT-4o-based single-agent systems generate roughly half as many (4.7--5.3).
System~M (GPT-4o), with its multi-agent architecture, generates concern volumes comparable to the Opus single-agent systems (${\sim}$10/paper).

\paragraph{Severity concentration.}
Systems~L and~A (Opus) mark over half their concerns as fatal or major (48--57\%), with near-identical severity profiles on accepted and rejected papers. 
On accepted papers, where our operational anchor assigns zero decisive blockers, this creates the high FDR documented in \S\ref{sec:discrimination}.
System~M shows the opposite pattern: only 25\% fatal+major on accepted papers, explaining its low FDR (0.10) but also its difficulty catching real blockers on rejected papers.

\paragraph{Verdict blindness.}
For most systems, concern count and severity distribution show minimal variation between accepted and rejected papers.
System~L (Opus) generates ${\sim}$11 concerns on accepted papers vs.\ ${\sim}$10 on rejected; the difference in fatal+major is 0.2 per paper.
System~A (GPT-4o) is nearly identical across verdicts (4.7 vs.\ 4.8 concerns, 2.9 vs.\ 2.9 fatal+major).
This quantifies the volume-without-discrimination pattern identified in
\S\ref{sec:discrimination}: the raw concern profiles confirm that these
systems do not modulate their output based on paper quality.

\section{Data Curation}
\label{app:curation}

\paragraph{Paper sourcing.}
Papers were sourced from OpenReview across ICLR~2026, NeurIPS~2025, and ICML~2025, filtered to a unified AI safety/alignment domain covering agent safety, alignment methods, red-teaming/jailbreaking, benchmarks, and human-AI oversight.
From an initial database of ${\sim}$2,950 papers in these topic areas, we selected 48 papers through a three-stage quality screening process.

\paragraph{Quality screening.}
\emph{Stage~1 (basic filters):} $\geq$3 substantive reviews, unambiguous AC decision with a meta-review that articulates the decisive factors, and complete OpenReview records (reviews, rebuttals, meta-review).
\emph{Stage~2 (extractability):} An LLM-as-judge pass assesses whether each paper's official reviews contain $\geq$2 specific, codable technical concerns (not just generic praise/criticism) and whether the meta-review provides enough reasoning to assign AC treatment labels.
\emph{Stage~3 (human overrides):} Borderline cases from Stage~2 are manually reviewed; papers with ambiguous decisions or low-quality reviews are excluded.

\paragraph{Composition.}
Table~\ref{tab:paper_composition} summarizes the 48-paper evaluation set.
All papers have $\geq$3 substantive reviews.
The set includes a mix of poster, oral, and spotlight acceptances and a range of rejection strengths.
We intentionally selected ``hard negatives'': 22 of 24 rejected papers have nontrivial scientific reasons for rejection (not desk rejects or formatting issues), ensuring that concern alignment metrics test genuine diagnostic ability.

\begin{table}[h]
\centering
\small
\caption{Evaluation set composition: 48 papers across 3 venues, balanced accept/reject, with topic and tier distribution.}
\label{tab:paper_composition}
\begin{tabular}{lcccc}
\toprule
\textbf{Venue} & \textbf{Accepted} & \textbf{Rejected} & \textbf{Total} & \textbf{Mean reviews} \\
\midrule
ICLR 2026 & 8 & 18 & 26 & 4.0 \\
NeurIPS 2025 & 10 & 5 & 15 & 3.9 \\
ICML 2025 & 6 & 1 & 7 & 3.6 \\
\midrule
\textbf{Total} & 24 & 24 & 48 & 3.9 \\
\bottomrule
\end{tabular}
\\[4pt]
\scriptsize
\textbf{Topics (non-exclusive):} safety (41), agent (32), alignment (22), benchmarks (19), attack/defense (8). \\
\textbf{Accepted tiers:} spotlight (10), poster (9), oral (5). \textbf{Concerns:} 670 official, 79 decisive blockers.
\end{table}

\paragraph{Version handling.}
Systems receive the camera-ready PDF for accepted papers and the original submission for rejected papers (6 rejected papers use the last revision when the submission PDF had rendering issues).
Concern extraction reads the full OpenReview record including rebuttals and meta-reviews, enabling AC treatment labels that distinguish decisive blockers from resolved concerns.

\paragraph{Sanitization.}
PDFs were sanitized to remove decision-revealing metadata (acceptance banners, camera-ready headers) via overlay redaction.
Post-fetch automated text extraction checks flagged papers with degraded rendering (broken fonts, garbled text); these were re-downloaded or excluded.
Paper titles, venues, and decisions are stored in a separate ground-truth file, not embedded in the PDFs provided to AI systems.

\subsection{Practical Lessons for Evaluation Infrastructure}
\label{app:rigor}

Concern-alignment evaluation requires stable, well-characterized paper
artifacts. Several data-quality issues arose during this study that we
document as practical guidance for future evaluations.

\paragraph{PDF integrity.}
Of the initial 48 PDFs downloaded from OpenReview, 15 had font-subsetting
rendering issues that left pages partially or fully unreadable. These were
detected by automated text-extraction checks and recovered by re-downloading
alternate versions. Three additional papers were flagged as broken by the
automated checker but were visually readable (false positives from reference
pages where margin line numbers dominated the text ratio). We recommend
combining automated quality checks with human visual inspection of a sample.

\paragraph{Decision-leaking metadata.}
Accepted papers from OpenReview may contain camera-ready headers, acceptance
banners, or author acknowledgments that reveal the decision. We developed a
dual-strategy sanitizer: overlay redaction (white rectangles) for benign
``Under review'' headers, and targeted tight-bbox redaction for
decision-leaking ``Published at'' or ``Accepted'' headers. An earlier
single-strategy approach corrupted font subsetting in 3 PDFs, illustrating
that sanitization itself can introduce artifacts.

\paragraph{Version mismatch.}
AI review systems receive the camera-ready PDF for accepted papers, which may incorporate post-rebuttal fixes.
We therefore track whether each resolved concern's fix is actually visible in
the reviewed PDF. Without this distinction, a system that correctly detects an
unresolved weakness can be penalized for ``re-escalating'' a concern that was
never actually fixed in the version it reviewed.

\paragraph{Rebuttal-process concerns.}
A small number of official decisive blockers concern the review process
itself (e.g., ``reviewers did not engage during discussion'') rather than
paper content. These are invisible to PDF-only systems and should be
excluded from detection denominators or flagged separately.

\section{Severity Determination Guidelines}
\label{app:severity_policy}

Severity assignment is the most subjective aspect of the concern alignment framework.
We make the rubric fully transparent so reviewers can calibrate independently.

\paragraph{Severity rubric.}
The extraction pipeline assigns severity based on three signals: structural cues in the review (section headers, explicit labels), language intensity, and available numeric scores.
The four levels are defined as follows:

\begin{itemize}[leftmargin=*,nosep]
\item \textbf{Fatal}: A fundamental flaw that, if true, invalidates the paper's central claim or makes the results uninterpretable. Examples: mathematical error in a core proof, data leakage between train and test, evaluation metric that does not measure what is claimed.
\item \textbf{Major}: A significant weakness that substantially undermines a central claim but does not invalidate the entire paper. The paper could potentially be revised to address it. Examples: missing critical baseline, evaluation on a single dataset when generalization is claimed, confound that could explain the main result.
\item \textbf{Moderate}: A meaningful gap that weakens but does not undermine the core contribution. Examples: limited ablation study, unclear methodology details that hinder reproducibility, moderate overclaiming relative to evidence.
\item \textbf{Minor}: A small issue that would improve the paper but does not affect the validity of the claims. Examples: notation inconsistencies, missing related work citations, presentation improvements.
\end{itemize}

\paragraph{Decisive flag assignment.}
A concern is marked \texttt{decisive=true} when the review context indicates it was intended as a reason to reject the paper.
Structural signals (appearing under ``reasons for rejection,'' co-occurring with low scores or an explicit reject decision) take priority; language intensity (``fundamental,'' ``invalidates,'' ``fatal flaw'') provides secondary evidence.
On accepted papers with a positive AC decision, the AC-aligned decisive-blocker count is zero by construction.

\paragraph{Representative examples.}
Table~\ref{tab:severity_examples} lists representative concerns from the evaluation set, showing how the rubric applies in practice.

\begin{table*}[t]
\centering
\small
\caption{Representative decisive vs.\ non-decisive concerns from the evaluation set. Reviewers can use these to calibrate their assessment of our severity assignments.}
\label{tab:severity_examples}
\begin{tabular}{p{0.48\textwidth}ccp{0.18\textwidth}}
\toprule
\textbf{Concern (paraphrased)} & \textbf{Severity} & \textbf{Decisive?} & \textbf{Rationale} \\
\midrule
\multicolumn{4}{l}{\emph{Decisive concerns (from rejected papers)}} \\
\addlinespace[2pt]
All harmful tasks contain explicit language, trivially filterable by input classifier & Fatal & Yes & Invalidates threat model \\
\addlinespace[2pt]
ASR metric misses end-to-end success; attack may not achieve real-world harm & Major & Yes & Undermines central evaluation \\
\addlinespace[2pt]
No comparison to adaptive defenses; only static baselines tested & Major & Yes & Missing critical baseline \\
\midrule
\multicolumn{4}{l}{\emph{Non-decisive concerns (from accepted papers)}} \\
\addlinespace[2pt]
Step-based evaluation may not capture end-to-end impact & Moderate & No & Acknowledged limitation, resolved in rebuttal \\
\addlinespace[2pt]
Limited to 4 web environments from WebArena & Moderate & No & Accepted limitation by AC \\
\addlinespace[2pt]
Notation inconsistency between Sections~2 and~4 & Minor & No & Presentation issue only \\
\addlinespace[2pt]
Missing comparison to concurrent work X & Minor & No & Would improve but does not invalidate \\
\bottomrule
\end{tabular}
\end{table*}

\section{Semantic Verification Calibration Exemplars}
\label{app:exemplars}

The independent auditor is calibrated with a fixed bank of 32 exemplars curated from externally audited edges across 5 audit rounds spanning 41 papers.
Each exemplar shows two concern summaries, the correct verdict, and a one-sentence reason; paper names are omitted to prevent anchoring.  Table~\ref{tab:exemplars} presents 8 representative exemplars spanning correct matches, near-misses, and all major error categories; the full bank of 32 is available in the supplementary materials.

\begin{table*}[t]
\centering
\scriptsize
\setlength{\tabcolsep}{3pt}
\caption{Representative calibration exemplars for semantic verification. Each row shows the official and agentic concern texts, the correct verdict, and a brief rationale.}
\label{tab:exemplars}
\begin{tabular}{@{}L{0.45cm}L{1.9cm}L{2.6cm}L{2.6cm}L{1.55cm}L{2.0cm}@{}}
\toprule
\textbf{ID} & \textbf{Category} & \textbf{Official concern} & \textbf{Agentic concern} & \textbf{Verdict} & \textbf{Rationale} \\
\midrule
E1 & Correct match & ``missing stronger baselines (VAE-based, trajectory auto-encoders)'' & ``only raw observations and outdated baseline compared, no modern methods'' & exact & Same gap, same scope \\
\addlinespace
E7 & Tricky correct & ``training data quality confound: no control for data filtering vs.\ streaming format'' & ``training data undergoes multi-stage filtering; unclear if gains come from data quality or method'' & partial & Same confound, different evidence \\
\addlinespace
E12 & Wrong match & ``extremely limited number of test scenes'' (data diversity) & ``narrow baseline set: only two methods compared'' (method diversity) & wrong match & More scenes $\neq$ more baselines \\
\addlinespace
E19 & Tricky wrong & ``using only one evaluation metric (e.g., ASR) is overly simplistic'' & ``no confidence intervals, significance tests, or multi-run reporting'' & wrong type (p$\to$r) & Metric choice $\neq$ reporting rigor \\
\addlinespace
E20 & Scope infl.\ (elab.) & ``validation metric is circular: uses model's own outputs as proxy'' & ``circular validation metric restated with additional confound language'' & wrong type (p$\to$e) & Same defect, more rhetoric \\
\addlinespace
E21 & Scope infl.\ (broad.) & ``no human evaluation of output realism'' & ``human validation needed: realism + correctness + safety audit'' & wrong type (e$\to$p) & Adds independent fix-actions \\
\addlinespace
E31 & Wrong severity & ``modest gains without statistical analysis [major]'' & ``missing statistical significance tests [moderate]'' & wrong sev. & Correct match; 2-level severity gap \\
\addlinespace
E32 & Cross-domain mismatch & ``missing analysis of \emph{why} the method works'' & ``missing comparison of output quality vs.\ alternatives'' & wrong match & Mechanism $\neq$ benchmark \\
\bottomrule
\end{tabular}
\end{table*}

\section{Extended Related Work}
\label{app:related_extended}

\paragraph{AI review generation and redesign.}
Existing AI-review systems cover several families: direct prompting \citep{liang2024can}, iterative reflection \citep{lu2024aiscientist}, structured progressive review \citep{oar2024}, multi-agent review generation \citep{darcy2024marg,gao2025reviewagents,zou2026diagpaper}, and specialized fine-tuned reviewer models such as OpenReviewer \citep{idahl2025openreviewer}.
These works primarily aim to generate better reviews or review comments.
Pairwise-comparison work \citep{zhang2025pairwise} asks a different systems question by replacing per-paper scoring with relative judgments across papers.
Concern alignment is orthogonal to all of them: it is an evaluation substrate that can audit any review-generating or review-ranking system as long as the system produces inspectable review artifacts.

\paragraph{Evaluating AI-generated reviews.}
\emph{Mind the Blind Spots} \citep{shin2025blindspots} operationalizes review focus as a distribution over predefined facets and compares LLM and human focus distributions.
\emph{ReviewEval} \citep{garg2025revieweval} scores AI reviews on holistic dimensions such as factuality, analytical depth, and constructiveness.
\emph{ReviewScore} \citep{ryu2025reviewscore} focuses on whether review points are misinformed by reconstructing explicit and implicit premises.
\emph{LimitGen} \citep{xu2025limitgen} asks whether models can identify critical limitations, and \emph{AAAR-1.0} \citep{lou2025aaar} benchmarks weakness identification as one research-assistance task.
\emph{DIAGPaper} \citep{zou2026diagpaper} is the closest generation-side paper on prioritization: it validates and ranks generated weaknesses for users.
Three aspects distinguish concern alignment from prior evaluation methods.
The unit of analysis is different (free-form concern instances rather than facets or holistic rubrics), the grounding is different (post-rebuttal AC treatment rather than review-to-review similarity alone), and the error model is different (explicit misses, phantoms, decision-weight errors, and resolved-concern re-escalation).

\paragraph{Fine-grained review structure.}
Prior work also decomposes peer review below the document level: argument pair extraction \citep{cheng2020ape}, discourse structure annotation \citep{kennard2022disapere}, rebuttal-effect analysis \citep{miyao2019does}, concern-decomposed rebuttal generation \citep{ma2026paper2rebuttal}, full-stage review--rebuttal datasets \citep{zhang2025re2}, and rebuttal-supervised actionable feedback generation \citep{wu2026rbtact}.
These resources motivate our choice to treat a review as a structured set of concern units rather than a monolithic text string, but they target generation, dataset construction, or discourse analysis rather than evaluation against decision rationale.

\paragraph{LLM use inside peer review.}
Observational and intervention studies document that LLMs already affect peer review practice \citep{liang2024monitoring,latona2024ailottery,thakkar2025reviewfeedback}.
That makes measurement design consequential: if AI reviews are going to be consumed by authors, reviewers, or future autonomous research agents, evaluation should reward not only overlap with human text but also calibrated prioritization and alignment with how decisions are actually made.

\section{Verdict-Stratified Accuracy}
\label{app:verdict_split}

Table~\ref{tab:verdict_split} reports verdict accuracy stratified by accepted and rejected papers.
These numbers reflect our extraction pipeline's inference of review tone (\S\ref{sec:setup}) and are sensitive to the inference method (Table~\ref{tab:verdict_sensitivity}).
They are included as context for the verdict inference audit below, not as standalone findings.

\begin{table}[h]
\centering
\small
\caption{Verdict-stratified accuracy (3-run mean $\pm$ std, pipeline inference). Several configurations show reject-heavy profiles with near-zero accepted-paper accuracy. These figures are method-dependent; see Table~\ref{tab:verdict_sensitivity} for sensitivity.}
\label{tab:verdict_split}
\scriptsize
\begin{tabular}{lcccc}
\toprule
\textbf{Sys.} & \textbf{Acc.\ acc.} & \textbf{Rej.\ acc.} & \textbf{Overall} & \textbf{Profile} \\
\midrule
L (Opus) & 2.8{\tiny$\pm$2.4} & 98.6{\tiny$\pm$2.4} & 50.7{\tiny$\pm$1.2} & Reject-heavy \\
A (Opus) & 8.3{\tiny$\pm$4.2} & 93.1{\tiny$\pm$4.8} & 50.7{\tiny$\pm$4.3} & Reject-heavy \\
O (Opus) & 34.7{\tiny$\pm$17} & 79.2{\tiny$\pm$0.0} & 56.9{\tiny$\pm$8.7} & Low-recall \\
L (GPT-4o) & 63.9{\tiny$\pm$2.4} & 51.4{\tiny$\pm$6.4} & 57.6{\tiny$\pm$3.2} & Moderate \\
A (GPT-4o) & 4.2{\tiny$\pm$4.2} & 97.2{\tiny$\pm$4.8} & 50.7{\tiny$\pm$1.2} & Reject-heavy \\
M (GPT-4o) & 79.2{\tiny$\pm$19} & 41.7{\tiny$\pm$18} & 60.4{\tiny$\pm$5.5} & High-var. \\
\bottomrule
\end{tabular}
\end{table}

\section{Verdict Inference Audit}
\label{app:verdict_audit}

Evaluating verdict accuracy requires an accept/reject label, but four of six configurations do not emit one; their reviews express accept-leaning or reject-leaning inclinations without a binary decision field.
The extraction pipeline therefore infers the implied recommendation from review tone, using a default-REJECT rule for ambiguous cases (\S\ref{sec:setup}).
To assess how this inference layer affects the paper's findings, we conducted a multi-method, multi-rater audit of all 288 reviews (48 papers $\times$ 6 configurations $\times$ run~1).

\paragraph{Three inference methods.}
(1)~\emph{Pipeline}: the existing extraction (Claude Sonnet, default-REJECT). (2)~\emph{Tone}: an independent rater reads the raw review without a default-REJECT instruction and assigns ACCEPT/REJECT/AMBIGUOUS. (3)~\emph{Gate}: an LLM classifies each major/fatal concern into gate categories (G1: claim--evidence mismatch, G2: baseline fairness, G4: validity, G5: novelty); deterministic rules produce REJECT if a fatal concern is present or $\geq$2 major/fatal concerns hit fundamental gates, ACCEPT if no fundamental triggers and a positive acceptance signal exists, AMBIGUOUS otherwise.
Neither the tone nor gate method is calibrated to the venue acceptance rate; they ask ``what does this review imply?'' rather than ``should this paper be accepted?''

\paragraph{Two independent raters.}
Rater~1 (Claude Opus) and Rater~2 (GPT-5.4 Pro) independently applied both the tone and gate methods to all 288 reviews, reading raw review text as the primary input and the extracted concern sheet as supplementary structure.
Inter-rater agreement on binary tone verdicts was 89.4\% ($\kappa = 0.77$); on binary gate verdicts, 93.8\% ($\kappa = 0.74$).
Both exceed the 0.60 threshold for substantial agreement.

\paragraph{Human adjudication.}
A human auditor reviewed 54 cases where the two raters disagreed on tone or gate verdict, or where both raters agreed that tone and gate methods diverged.
For each case, the auditor read the full review text, both raters' reasoning, and assigned a final verdict.
Of 25 binary-verdict audited cases, the human agreed with the gate analysis 84\% of the time (both raters), compared to 64\% for Rater~1's tone and 28\% for Rater~2's tone.
All 48 System~M reviews were flagged as structurally unreliable due to multi-agent coordination artifacts.

\paragraph{Accepted-paper accuracy under alternative methods.}
Table~\ref{tab:verdict_sensitivity} shows accepted-paper accuracy for each configuration under all methods plus the human-adjudicated resolution.

\begin{table}[h]
\small
\centering
\caption{Accepted-paper accuracy (\%) under five verdict inference approaches and a resolved column, for run~1 of each configuration (the audit covered 48 papers $\times$ 6 configurations $\times$ run~1 = 288 reviews).
Pipeline values here are single-run and may differ from the 3-run means reported in the main text.
The Resolved column uses human adjudication for the 54 audited disagreement cases and rater-consensus resolution elsewhere.
System~M is flagged ($\dagger$): all reviews structurally unreliable.}
\label{tab:verdict_sensitivity}
\begin{tabular}{lcccccc}
\toprule
\textbf{Config} & \textbf{Pipe.} & \textbf{Tone\textsubscript{R1}} & \textbf{Tone\textsubscript{R2}} & \textbf{Gate\textsubscript{R1}} & \textbf{Gate\textsubscript{R2}} & \textbf{Resolved} \\
\midrule
L (Opus) & 0 & 4 & 42 & 0 & 0 & 0 \\
L (GPT-4o) & 62 & 100 & 92 & 46 & 50 & 83 \\
A (Opus) & 4 & 4 & 4 & 0 & 0 & 0 \\
A (GPT-4o) & 0 & 0 & 0 & 0 & 12 & 0 \\
O (Opus) & 29 & 29 & 50 & 4 & 12 & 33 \\
M (GPT-4o)$^\dagger$ & 58 & 58 & 67 & 8 & 38 & 33 \\
\bottomrule
\end{tabular}
\end{table}

\paragraph{Claim sensitivity.}
A~(Opus) shows a consistent reject-heavy profile across all methods and both raters (0--4\%).
L~(Opus) is similarly reject-heavy under most methods ($\leq$\,4\%), though Rater~2's tone reading is an outlier at 42\%; the human-adjudicated resolution and all other method/rater combinations agree on $\leq$\,4\%.
The model-effect swing (L~Opus $\to$ L~GPT-4o) ranges from 46pp (gate) to 96pp (tone), with the pipeline value falling within this range.
System~M's pipeline accuracy drops sharply under gate-based inference and human adjudication, while tone-based methods remain nearer to pipeline levels, consistent with structural artifacts that make accept/reject intent hard to recover reliably (Table~\ref{tab:verdict_sensitivity}).
All concern-level diagnostics reported in the paper---recall, FDR, decisive precision, phantom rates, attention profiles, ICC, and top-$K$ analyses---are independent of the verdict inference method because they are computed or stratified by official verdict rather than predicted verdict.

\paragraph{Root observation.}
When a review does not make its recommendation explicit, accept/reject must be inferred from tone or concern structure, and that inference can be unstable.
In this pilot, verdict ambiguity and calibration failures sometimes co-occur, but the six audited configurations do not show a simple monotone relationship.
Concern-level evaluation is useful because it provides stable diagnostics even when verdict extraction is noisy.

\section*{Disclosure of LLM Use}
This work used large language models for parts of the evaluation pipeline reported here, for preparation of some plots and analysis artifacts, and for limited assistance with manuscript drafting and revision. All such outputs were reviewed and verified by the author, who takes responsibility for all claims, analyses, and conclusions.
\end{document}

%% file: case_studies/paper_d_spard_compare.tex
\begin{table*}[t]
\centering\footnotesize
\caption{Annotated concern comparison for Paper~D (REJECTED) --- \textit{Review quality gap.} Official concerns (left) matched against 2 systems. Badges show match type, severity alignment, and AC treatment. System~A catches all 3 content-related decisive blockers; O15 is a process concern invisible to PDF-only systems.}
\label{tab:casestudy_d}
\begin{tabular}{p{0.28\textwidth}p{0.33\textwidth}p{0.33\textwidth}}
\toprule
\textbf{Official concern} & \textbf{System~A (Opus)} & \textbf{System~O (Opus)} \\
\midrule
\textbf{O1}: The core technical components are extensions of existing paradigms\ldots \newline \badge{phantomred}{fatal} \badge{phantomred}{decisive blocker} \badge{decisiveorange}{DECISIVE} & \badge{matchgreen}{exact} \badge{matchgreen}{ok} \newline \textit{Limited novelty --- the method uses standard projected gradient desc\ldots} & \textcolor{gray!50}{no match} \\[4pt]
\textbf{O2}: The paper compares against only a small number of defense methods, omitting rece\ldots \newline \badge{phantomred}{fatal} \badge{phantomred}{decisive blocker} \badge{decisiveorange}{DECISIVE} & \badge{matchgreen!60!black}{partial} \badge{matchgreen}{ok} \newline \textit{Incomplete baseline comparisons with recent defense methods} & \textcolor{gray!50}{no match} \\[4pt]
\textbf{O3}: The defense is validated on only two LLM architectures (Qwen-2.5-7B, LLaMA-3.2-3\ldots \newline \badge{decisiveorange}{major} \badge{phantomred}{decisive blocker} \badge{decisiveorange}{DECISIVE} & \badge{matchgreen!60!black}{partial} \badge{matchgreen}{ok} \newline \textit{Limited experimental scope undermines generalizability claim\ldots} & \textcolor{gray!50}{no match} \\[4pt]
\textit{O15 (process concern, excluded)}: Reviewers did not engage during discussion\ldots \newline \badge{gray}{major} \badge{gray}{process-only} \badge{gray}{undetectable from PDF} & \textcolor{gray!50}{---} & \textcolor{gray!50}{---} \\[4pt]
\textbf{O4}: The paper does not clearly explain how 'informative' and 'mutually diverse' are \ldots \newline \badge{decisiveorange}{major} \badge{gray}{not mentioned}  & \textcolor{gray!50}{no match} & \textcolor{gray!50}{no match} \\[4pt]
\midrule
\textit{Phantoms (unmatched agentic)} & 6 phantoms \newline \scriptsize{A02, A03, A04} & 7 phantoms \newline \scriptsize{A1, A2, A3} \\
\bottomrule
\end{tabular}
\end{table*}

%% file: case_studies/paper_a_hexgoedel_compare.tex
\begin{table*}[t]
\centering\footnotesize
\caption{Annotated concern comparison for Paper~A (ACCEPTED) --- \textit{Model effect on calibration.} Same method (System~L), different model. Opus marks all 6 concerns decisive (FDR = 100\%); GPT-4o marks 3 of 7 (FDR = 43\%).\textsuperscript{\textdagger} Opus rejects; GPT-4o accepts. {\footnotesize \textsuperscript{\textdagger}Per-paper FDR; aggregate FDR across all accepted papers differs.}}
\label{tab:casestudy_a}
\begin{tabular}{p{0.28\textwidth}p{0.33\textwidth}p{0.33\textwidth}}
\toprule
\textbf{Official concern} & \textbf{System~L (Opus)} & \textbf{System~L (GPT-4o)} \\
\midrule
\textbf{O6}: Core theoretical result is near-tautological under stated assumptions\ldots \newline \badge{decisiveorange}{major} \badge{gray}{acc.\ limit.}  & \badge{matchgreen}{exact} \badge{matchgreen}{ok} \newline \textit{Theorem proof reduces to standard result once key assumption granted} \newline \badge{decisiveorange}{DECISIVE} & \badge{matchgreen!60!black}{partial} \badge{matchgreen}{ok} \newline \textit{Theoretical framework is difficult to grasp; proof accessibility\ldots} \\[4pt]
\textbf{O12}: Performance metric is a probabilistic estimate; predictive accuracy uncertain\ldots \newline \badge{decisiveorange}{moderate} \badge{rebuttalpurple}{resolved}  & \badge{matchgreen!60!black}{partial} \badge{matchgreen}{ok} \newline \textit{Metric estimator validity may be a structural artifact\ldots} \newline \badge{decisiveorange}{DECISIVE} & \badge{gray}{related} \badge{gray}{n/a} \newline \textit{Estimation process not clearly articulated\ldots} \\[4pt]
\textbf{O4}: Evaluation limited to coding benchmarks; generalization unverified\ldots \newline \badge{decisiveorange}{major} \badge{rebuttalpurple}{resolved}  & \badge{gray}{related} \badge{gray}{n/a} \newline \textit{Generalization experiment confounds dataset and backbone\ldots} \newline \badge{decisiveorange}{DECISIVE} & \badge{matchgreen}{exact} \badge{matchgreen}{ok} \newline \textit{Evaluation limited to two coding benchmarks\ldots} \newline \badge{decisiveorange}{DECISIVE} \\[4pt]
\textbf{O7}: Algorithm hard to follow; prompts not included; baseline parity unclear\ldots \newline \badge{decisiveorange}{moderate} \badge{rebuttalpurple}{resolved} \textit{(rebuttal-only)}  & \textcolor{gray!50}{no match} & \badge{matchgreen!60!black}{partial} \badge{matchgreen}{ok} \newline \textit{Lacks detailed instructions for reproduction\ldots} \\[4pt]
\textbf{O9}: Ethical risks of recursive self-improvement undiscussed\ldots \newline \badge{decisiveorange}{moderate} \badge{rebuttalpurple}{resolved}  & \textcolor{gray!50}{no match} & \badge{matchgreen!60!black}{partial} \badge{matchgreen}{ok} \newline \textit{Limitations section incomplete; broader impact missing\ldots} \\[4pt]
\midrule
\textit{Phantoms (unmatched agentic)} & 2 phantoms \newline \scriptsize{A03: statistical rigor (single run, no CIs)} \newline \scriptsize{A04: no hyperparameter ablation} \newline \scriptsize{Both marked \badge{decisiveorange}{DECISIVE}} & 1 phantom \newline \scriptsize{A5: theoretical assumptions may not hold} \\
\bottomrule
\end{tabular}
\end{table*}

%% file: case_studies/paper_h_decoupling_compare.tex
\begin{table*}[t]
\centering\footnotesize
\caption{Annotated concern comparison for Paper~H (REJECTED) --- \textit{Maximum-contrast model effect.} Official concerns (left) matched against 2 systems. Badges show match type, severity alignment, and AC treatment.}
\label{tab:casestudy_h}
\begin{tabular}{p{0.28\textwidth}p{0.33\textwidth}p{0.33\textwidth}}
\toprule
\textbf{Official concern} & \textbf{System~L (Opus)} & \textbf{System~L (GPT-4o)} \\
\midrule
\textbf{O1}: The paper's core contribution is a repackaging of existing fine-tuning techniques with a safety-preservation objective\ldots \newline \badge{phantomred}{fatal} \badge{phantomred}{decisive blocker} \badge{decisiveorange}{DECISIVE} & \badge{matchgreen}{exact} \badge{matchgreen}{ok} \newline \textit{The core contribution is a repackaging of existing fine-tuning techniques with a safety regularizer\ldots} & \textcolor{gray!50}{no match} \\[4pt]
\textbf{O2}: The theoretical framework is largely heuristic: claims about subspace separation lack formal proof\ldots \newline \badge{phantomred}{fatal} \badge{phantomred}{decisive blocker} \badge{decisiveorange}{DECISIVE} & \badge{matchgreen}{exact} \badge{matchgreen}{ok} \newline \textit{The theoretical framework is largely heuristic, lacking formal proof of the claimed separation\ldots} & \badge{gray}{related} \badge{gray}{n/a} \newline \textit{Weakness in theoretical framework} \\[4pt]
\textbf{O11}: The key analytical result --- that the safety-related weights occupy a distinct subspace --- is a trivial consequence of the decomposition method\ldots \newline \badge{decisiveorange}{major} \badge{phantomred}{decisive blocker} \badge{decisiveorange}{DECISIVE} & \badge{matchgreen}{exact} \badge{matchgreen}{ok} \newline \textit{The decomposition-based finding is a trivial consequence of the low-rank structure\ldots} & \textcolor{gray!50}{no match} \\[4pt]
\textbf{O3}: The normalization procedure is mathematically incorrect: dividing\ldots \newline \badge{decisiveorange}{major} \badge{rebuttalpurple}{resolved}  & \textcolor{gray!50}{no match} & \textcolor{gray!50}{no match} \\[4pt]
\textbf{O4}: Missing comparison to closely related prior methods that employ similar safety-preserving fine-tuning\ldots \newline \badge{decisiveorange}{major} \badge{decisiveorange}{unresolved}  & \badge{matchgreen!60!black}{partial} \badge{matchgreen}{ok} \newline \textit{Missing comparison against closely related safety-preserving\ldots} & \badge{matchgreen!60!black}{partial} \badge{matchgreen}{ok} \newline \textit{Incomplete comparison to related safety alignment methods} \\[4pt]
\midrule
\textit{Phantoms (unmatched agentic)} & 6 phantoms \newline \scriptsize{A5, A7, A8} & 1 phantoms \newline \scriptsize{A4} \\
\bottomrule
\end{tabular}
\end{table*}

%% file: case_studies/paper_c_aria_compare.tex
\begin{table*}[t]
\centering\footnotesize
\caption{Annotated concern comparison for Paper~C (ACCEPTED) --- \textit{Harmful vs.\ benign phantoms.} Official concerns (left) matched against 2 systems. Badges show match type, severity alignment, and AC treatment.}
\label{tab:casestudy_c}
\begin{tabular}{p{0.28\textwidth}p{0.33\textwidth}p{0.33\textwidth}}
\toprule
\textbf{Official concern} & \textbf{System~O (Opus)} & \textbf{System~L (Opus)} \\
\midrule
\textbf{O11}: The main theorem depends on a very strong bisimulation assumption, m\ldots \newline \badge{decisiveorange}{major} \badge{gray}{acc.\ limit.}  & \badge{matchgreen!60!black}{partial} \badge{matchgreen}{ok} \newline \textit{The core theoretical framework has unaddressed assumptions\ldots} & \badge{matchgreen}{exact} \badge{matchgreen}{ok} \newline \textit{The bisimulation assumption is strong and \ldots} \\[4pt]
\textbf{O1}: No evaluation of the online method in adversarial (multi-agent) tasks, limiting eviden\ldots \newline \badge{gray}{moderate} \badge{rebuttalpurple}{resolved}  & \textcolor{gray!50}{no match} & \badge{matchgreen!60!black}{partial} \badge{matchgreen}{ok} \newline \textit{Evaluation scope is too narrow, missing adversarial/complex \ldots} \\[4pt]
\textbf{O2}: Skepticism about the method's generalization to more complex negotiation/dialogue task\ldots \newline \badge{gray}{moderate} \badge{rebuttalpurple}{resolved}  & \textcolor{gray!50}{no match} & \textcolor{gray!50}{no match} \\[4pt]
\textbf{O3}: The clustering process assumes discrete intention boundaries, which may not hold\ldots \newline \badge{gray}{moderate} \badge{gray}{acc.\ limit.}  & \textcolor{gray!50}{no match} & \badge{gray}{related} \badge{gray}{n/a} \newline \textit{Clustering assumption limitations in complex settings} \\[4pt]
\textbf{O4}: Method relies heavily on embedding quality with limited discussion of how embedd\ldots \newline \badge{gray}{moderate} \badge{rebuttalpurple}{resolved}  & \textcolor{gray!50}{no match} & \badge{matchgreen}{exact} \badge{matchgreen}{ok} \newline \textit{Performance is sensitive to embedding model choice} \\[4pt]
\midrule
\textit{Phantoms (unmatched agentic)} & 7 phantoms \newline \scriptsize{A1: ``Proof of a supplementary lemma contains an incorrect algebraic decomposition; counterexample: Var=380.5 vs 37.5''} \badge{phantomred}{fatal} \newline \scriptsize{+ 6 others (A2, A3, \ldots)} & 6 phantoms \newline \scriptsize{A2, A3, A4} \\
\bottomrule
\end{tabular}
\end{table*}

%% file: case_studies/paper_e_osharm_compare.tex
\begin{table*}[t]
\centering\footnotesize
\caption{Annotated concern comparison for Paper~E (ACCEPTED) --- \textit{FDR metric validation.} Official concerns (left) matched against 2 systems. Badges show match type, severity alignment, and AC treatment.}
\label{tab:casestudy_e}
\begin{tabular}{p{0.28\textwidth}p{0.33\textwidth}p{0.33\textwidth}}
\toprule
\textbf{Official concern} & \textbf{System~L (Opus)} & \textbf{System~A (Opus)} \\
\midrule
\textbf{O1}: The LLM judge used for automated safety evaluation has very low agreement with h\ldots \newline \badge{decisiveorange}{major} \badge{rebuttalpurple}{resolved}  & \badge{matchgreen}{exact} \badge{matchgreen}{ok} \newline \textit{LLM judge has insufficient reliability for automated safety \ldots} & \badge{matchgreen}{exact} \badge{matchgreen}{ok} \newline \textit{LLM judge has low reliability for automated safety evaluatio\ldots} \\[4pt]
\textbf{O2}: The benchmark is limited to a single OS environment; different operating systems have dif\ldots \newline \badge{gray}{moderate} \badge{gray}{acc.\ limit.}  & \textcolor{gray!50}{no match} & \textcolor{gray!50}{no match} \\[4pt]
\textbf{O3}: The paper does not address concerns about benchmark gaming: models could be trai\ldots \newline \badge{gray}{moderate} \badge{gray}{not mentioned}  & \badge{gray}{related} \badge{gray}{n/a} \newline \textit{Limited benchmark size raises concerns} & \badge{matchgreen!60!black}{partial} \badge{matchgreen}{ok} \newline \textit{Small benchmark scale limits reliability} \\[4pt]
\textbf{O5}: Unclear how the simulation-based evaluation framework informs real-world risk: d\ldots \newline \badge{gray}{moderate} \badge{gray}{acc.\ limit.}  & \badge{gray}{related} \badge{gray}{n/a} \newline \textit{What benchmark results actually tell us about agent safety} & \badge{matchgreen!60!black}{partial} \badge{matchgreen}{ok} \newline \textit{Gap between benchmark simulation and real-world deployment s\ldots} \\[4pt]
\midrule
\textit{Phantoms (unmatched agentic)} & 13 phantoms \newline \scriptsize{A1, A2, A3} & 8 phantoms \newline \scriptsize{A3, A4, A5} \\
\bottomrule
\end{tabular}
\end{table*}

%% file: case_studies/paper_f_vpibench_compare.tex
\begin{table*}[t]
\centering\footnotesize
\caption{Annotated concern comparison for Paper~F (ACCEPTED) --- \textit{High recall, wrong decision weight.} Official concerns (left) matched against 2 systems. Badges show match type, severity alignment, and AC treatment.}
\label{tab:casestudy_f}
\begin{tabular}{p{0.28\textwidth}p{0.33\textwidth}p{0.33\textwidth}}
\toprule
\textbf{Official concern} & \textbf{System~A (Opus)} & \textbf{System~L (Opus)} \\
\midrule
\textbf{O1}: The proposed attacks lack a clear and significant distinction from prior work\ldots \newline \badge{decisiveorange}{major} \badge{rebuttalpurple}{resolved}  & \badge{matchgreen!60!black}{partial} \badge{matchgreen}{ok} \newline \textit{Novelty concerns related to overlap with prior attack methods\ldots} & \badge{matchgreen}{exact} \badge{matchgreen}{ok} \newline \textit{The proposed attacks lack novelty --- essentially repurposing\ldots} \\[4pt]
\textbf{O2}: The paper's conceptual novelty is limited, primarily packaging known ideas without n\ldots \newline \badge{decisiveorange}{major} \badge{gray}{acc.\ limit.}  & \badge{matchgreen}{exact} \badge{matchgreen}{ok} \newline \textit{Limited conceptual novelty --- packages known attack techniques\ldots} & \badge{matchgreen!60!black}{partial} \badge{matchgreen}{ok} \newline \textit{Limited conceptual novelty --- packaging known ideas without n\ldots} \\[4pt]
\textbf{O3}: The reliance on an ensemble of LLMs as judges for evaluating agent behavior requ\ldots \newline \badge{decisiveorange}{major} \badge{rebuttalpurple}{resolved}  & \badge{matchgreen}{exact} \badge{decisiveorange}{under} \newline \textit{LLM-as-judge evaluation not sufficiently validated by human \ldots} & \textcolor{gray!50}{no match} \\[4pt]
\textbf{O8}: The threat model seems unrealistic in assuming that large platforms such as BBC \ldots \newline \badge{decisiveorange}{major} \badge{rebuttalpurple}{resolved}  & \badge{matchgreen!60!black}{partial} \badge{matchgreen}{ok} \newline \textit{Threat model realism concerns --- gap between evaluation setup\ldots} & \badge{matchgreen}{exact} \badge{matchgreen}{ok} \newline \textit{Threat model is unrealistic --- assumes attacker can fully com\ldots} \\[4pt]
\textbf{O4}: The benchmark does not include agents that have undergone extensive task-specifi\ldots \newline \badge{gray}{moderate} \badge{gray}{acc.\ limit.}  & \textcolor{gray!50}{no match} & \textcolor{gray!50}{no match} \\[4pt]
\midrule
\textit{Phantoms (unmatched agentic)} & 2 phantoms \newline \scriptsize{A2, A3} & 1 phantom \newline \scriptsize{A05} \\
\bottomrule
\end{tabular}
\end{table*}

%% file: case_studies/paper_g_rebench_compare.tex
\begin{table*}[t]
\centering\footnotesize
\caption{Annotated concern comparison for Paper~G (ACCEPTED) --- \textit{Different verdict, different grounding.} Official concerns (left) matched against 2 systems. Badges show match type, severity alignment, and AC treatment.}
\label{tab:casestudy_g}
\begin{tabular}{p{0.28\textwidth}p{0.33\textwidth}p{0.33\textwidth}}
\toprule
\textbf{Official concern} & \textbf{System~L (Opus)} & \textbf{System~O (Opus)} \\
\midrule
\textbf{O1}: Unlimited access to the scoring function gives an unfair advantage to fast-itera\ldots \newline \badge{decisiveorange}{major} \badge{gray}{acc.\ limit.}  & \badge{gray}{related} \badge{gray}{n/a} \newline \textit{Structural advantages for AI agents undermine comparison fai\ldots} & \textcolor{gray!50}{no match} \\[4pt]
\textbf{O3}: The benchmark tasks are relatively small-scale (completable in one day to one we\ldots \newline \badge{decisiveorange}{major} \badge{gray}{not mentioned}  & \badge{matchgreen}{exact} \badge{matchgreen}{ok} \newline \textit{Benchmark tasks are too small-scale and self-contained to ge\ldots} & \textcolor{gray!50}{no match} \\[4pt]
\textbf{O2}: Claiming the environments evaluate open-ended ML research engineering overclaims\ldots \newline \badge{gray}{moderate} \badge{gray}{acc.\ limit.}  & \badge{matchgreen!60!black}{partial} \badge{matchgreen}{ok} \newline \textit{Paper overclaims what the benchmark measures relative to rea\ldots} & \textcolor{gray!50}{no match} \\[4pt]
\textbf{O4}: The benchmark only measures full automation; if an AI agent can do 90\% of the wo\ldots \newline \badge{gray}{moderate} \badge{gray}{not mentioned}  & \textcolor{gray!50}{no match} & \textcolor{gray!50}{no match} \\[4pt]
\textbf{O5}: The two scaffolds tested are somewhat arbitrary, and results show \ldots \newline \badge{gray}{moderate} \badge{gray}{not mentioned}  & \textcolor{gray!50}{no match} & \textcolor{gray!50}{no match} \\[4pt]
\midrule
\textit{Phantoms (unmatched agentic)} & 3 phantoms \newline \scriptsize{A5, A6, A7} & 6 phantoms \newline \scriptsize{A1, A2, A3} \\
\bottomrule
\end{tabular}
\end{table*}